%% file: main.tex
\documentclass{article}

\usepackage{microtype}
\usepackage{graphicx}
\usepackage{subfigure}
\usepackage{booktabs} 

\usepackage[pagebackref=true]{hyperref}

\usepackage[accepted]{icml2023}

\usepackage{amsmath}
\usepackage{amssymb}
\usepackage{mathtools}
\usepackage{amsthm}

\usepackage[capitalize,noabbrev]{cleveref}

\usepackage[textsize=tiny]{todonotes}

\usepackage{amsmath,amsthm,amssymb,amscd}

\usepackage[ruled,vlined, algo2e]{algorithm2e}

\usepackage{enumerate}
\usepackage{fancyhdr}
\usepackage{mathrsfs}
\usepackage{xcolor}
\usepackage{graphicx}
\usepackage{listings}
\usepackage{hyperref}
\usepackage{caption}
\usepackage{adjustbox}
\usepackage{multirow}
\usepackage{microtype}
\usepackage{tikz}
\usepackage{pgfplots}
\pgfplotsset{compat=newest}
\usepgfplotslibrary{groupplots}
\usepgfplotslibrary{dateplot}
\usepackage{caption}
\usepackage{todonotes}
\usepackage{enumitem}
\tikzset{
  causalvar/.style      = {draw, circle, node distance = 2cm}
}

\usepackage{amsmath}
\usetikzlibrary{matrix}
\usetikzlibrary{shapes,arrows}
\usepackage{mathtools}
\usepackage{bbm}
\DeclareMathOperator*{\argmin}{arg\,min}
\DeclareMathOperator*{\argmax}{arg\,max}
\DeclareMathOperator{\E}{\mathbb{E}}
\newcommand{\cH}{\mathcal{H}}
\newcommand{\cX}{\mathcal{X}}
\newcommand{\cP}{\mathcal{P}}
\newcommand{\cA}{\mathcal{A}}

\newcommand{\cY}{\mathcal{Y}}

\newcommand{\cS}{\mathcal{S}}
\newcommand{\cN}{\mathcal{N}}
\newcommand{\cL}{\mathcal{L}}

\newcommand{\cF}{\mathcal{F}}

\renewcommand{\epsilon}{\varepsilon}

\newcommand{\R}{\mathbb{R}}
\newcommand{\defeq}{\vcentcolon=}

\usepackage{bbold}

\newcommand{\notindep}{\not\!\perp\!\!\!\perp}
\newcommand{\indep}{\perp \!\!\! \perp}

\newtheorem{exmp}{Example}[section]

\newtheorem{proposition}{Proposition}[section]

\newtheorem{definition}{Definition}[section]
\newtheorem{assum}{Assumption}[section]
\usepackage{tabularx}

\usepackage{amsmath}
\newcommand*\diff{\mathop{}\!\mathrm{d}}

\usepackage{microtype}
\usepackage{graphicx}
\usepackage{subfigure}
\usepackage{booktabs} 
\PassOptionsToPackage{numbers}{natbib}
\usepackage{natbib}

\usepackage{amsmath,amsthm,amssymb,amscd}

\usepackage{enumerate}
\usepackage{fancyhdr}
\usepackage{mathrsfs}
\usepackage{xcolor}
\usepackage{graphicx}
\usepackage{listings}
\usepackage{caption}
\usepackage{adjustbox}
\usepackage{multirow}
\usepackage{microtype}
\usepackage{tikz}
\usepackage{pgfplots}
\pgfplotsset{compat=newest}
\usepgfplotslibrary{groupplots}
\usepgfplotslibrary{dateplot}
\usepackage{caption}
\usepackage{todonotes}
\tikzset{
  causalvar/.style      = {draw, circle, node distance = 2cm}
}
\usetikzlibrary{matrix}
\usetikzlibrary{shapes,arrows}
\usepackage{mathtools}
\usepackage{dsfont}

\usepackage{amssymb}
\usepackage{pifont}

 \newtheorem{innercustomprop}{Proposition}
\newenvironment{customprop}[1]
  {\renewcommand\theinnercustomprop{#1}\innercustomprop}
  {\endinnercustomprop}

\newcommand{\Var}{\mathrm{Var}}
\newcommand{\Cov}{\mathrm{Cov}}

\newcommand{\benchpricing}[0]{\emph{Pricing}}
\newcommand{\benchpotential}[0]{\emph{Advertising}}

\newcommand{\benchscene}[0]{\emph{Scene}}
\newcommand{\benchyeast}[0]{\emph{Yeast}}
\newcommand{\benchtmc}[0]{\emph{TMC2007}}

\icmltitlerunning{Sequential Counterfactual Risk Minimization}

\begin{document}

\twocolumn[
\icmltitle{Sequential Counterfactual Risk Minimization}



\icmlsetsymbol{equal}{*}

\begin{icmlauthorlist}
\icmlauthor{Houssam Zenati}{1,2}
\icmlauthor{Eustache Diemert}{1}
\icmlauthor{Matthieu Martin}{1}
\icmlauthor{Julien Mairal}{2}
\icmlauthor{Pierre Gaillard}{2}

\end{icmlauthorlist}

\icmlaffiliation{1}{Criteo AI Lab}
\icmlaffiliation{2}{Univ. Grenoble Alpes, Inria, CNRS, Grenoble INP, LJK, 38000 Grenoble, France}

\icmlcorrespondingauthor{Houssam Zenati}{housszenati@gmail.com}

\icmlkeywords{Machine Learning, ICML}

\vskip 0.3in
]



\printAffiliationsAndNotice{} 

\begin{abstract}
Counterfactual Risk Minimization (CRM) is a framework for dealing with the logged bandit feedback problem, where the goal is to improve a logging policy using offline data. In this paper, we explore the case where it is possible to deploy learned policies multiple times and acquire new data. We extend the CRM principle and its theory to this scenario, which we call "Sequential Counterfactual Risk Minimization (SCRM)." We introduce a novel counterfactual estimator and identify conditions that can improve the performance of CRM in terms of excess risk and regret rates, by using an analysis similar to restart strategies in accelerated optimization methods. We also provide an empirical evaluation of our method in both discrete and continuous action settings, and demonstrate the benefits of multiple deployments of CRM.


\end{abstract}
\section{Introduction}
\label{introduction} 
Counterfactual reasoning in the logged bandit problem has become a common task for practitioners in a wide range of applications such as recommender systems \citep{swaminathan2012}, ad placements \citep{bottou2012} or precision medicine \citep{Kallus2018PolicyEA}. Such a task typically consists in learning an optimal decision policy from logged contextual features and partial feedbacks induced by predictions from a logging policy. To do so, the logged data is originally obtained from a randomized data collection experiment. However, the success of counterfactual risk minimization is  highly dependent on the quality of the logging policy and its ability to sample meaningful actions. 

Counterfactual reasoning can be challenging due to large variance issues associated with counterfactual estimators \citep{swaminathan2015}. Additionally, as pointed out by \citet{bottou2012}, confidence intervals obtained from counterfactual estimates may not be sufficiently accurate to select a final policy from offline data \citep{coindice2020}. This can occur when the logging policy does not sufficiently explore the action space. To address this, one option is to simply collect additional data from the same logging system to increase the sample size. However, it may be more efficient to use already collected data to design a better data collection experiment through a sequential design approach \citep[see Section 6.4]{bottou2012}. It is thus appealing to consider successive policy deployments when possible.


We tackle this sequential design problem and are interested in multiple deployments of the CRM setup of \citet{swaminathan2012}, which we call sequential counterfactual risk minimization (SCRM).
SCRM performs a sequence of data collection experiments by determining at each round a policy using data samples collected during previous experiments. The obtained policy is then deployed for the next round to collect additional samples. Such a sequential decision making system thus entails designing an adaptive learning strategy that minimizes the excess risk and expected regret of the learner. In contrast to the conservative learning strategy in CRM, the exploration induced by sequential deployments of enhanced logging policies should allow for improved excess risk and regret guarantees. Yet, obtaining such guarantees is nontrivial and we address it in this work.

In order to accomplish this, we first propose a new counterfactual estimator that controls the variance and analyze its convergence guarantees. Specifically, we obtain an improved dependence on the variance of importance weights between the optimal and logging policy. Second, leveraging this estimator and a weak assumption on the concentration of this variance term, we show how the error bound sequentially concentrates through CRM rollouts. This allows us to improve the excess risk bounds convergence rate as well as the regret rate. Our analysis employs methods similar to restart strategies in acceleration methods \citep{nesterov2012GradientMF} and optimization for strongly convex functions \citep{boyd_vandenberghe_2004}. We also conduct numerical experiments to demonstrate the effectiveness of our method in both discrete and continuous action settings, and how it improves upon CRM and other existing methods in the literature.



\section{Related Work}
\input{related_work.tex}

\section{Sequential Counterfactual Risk Minimization}

In this section, we introduce the (CRM) framework and motivate the use of sequential designs for (SCRM).

\paragraph{Notations} For random variables $x \sim \cP_\cX$, $a \sim \pi_{\theta}(\cdot | x)$ and $y \sim \mathcal P_\mathcal{Y}(\cdot | x, a)$, we write the expectation $\E_{x, \theta, y} [\cdot]~=~\E_{x \sim \cP_\cX, a \sim \pi_{\theta}(\cdot | x), y \sim \mathcal P_\mathcal{Y}(\cdot | x, a)} [\cdot]$ and do the same for the variance $\Var_{x, \theta, y}$. Moreover, $\lesssim$ denotes approximate inequalities up to universal multiplicative terms.

\subsection{Background}

In the counterfactual risk minimization (CRM) problem, we are given $n$ logged observations $(x_i, a_i, y_i)_{i = 1, \ldots, n}$ where contexts $x_i \in \cX$ are sampled from a stochastic environment distribution $x_i \sim \cP_\cX$, actions $a_i \sim \pi_{\theta_0}(\cdot | x_i)$ are drawn from a logging policy $\pi_{\theta_0}$ with a model $\theta_0$ in a parameter space $\Theta$. The losses are drawn from a conditional distribution $y_i \sim \mathcal P_\mathcal{Y}(\cdot | x_i, a_i)$. We note $\pi_{0,i}=\pi_{\theta_0}(a_i | x_i)$ the associated propensities and assume them to be known. We will assume that the policies in $\pi_\theta, \theta \in \Theta$ admit densities so that the propensities will denote the density function of the logging policy on the actions given the contexts. The expected risk of a model $\theta$ is defined as:
\begin{equation}
    L(\theta) = \displaystyle \E_{x, \theta, y} \left[ y \right].
    \label{eq:expected_risk}
\end{equation}
Counterfactual reasoning uses the logged data sampled from the logging policy associated to $\theta_0$ to estimate the risk of any model $\theta \in \Theta$ with importance sampling:
\begin{equation}
   L(\theta) = \displaystyle \E_{x, \theta_0, y} \left[ y \dfrac{\pi_{\theta}(a |x)}{\pi_{\theta_0}(a |x)} \right],
    \label{eq:importancesampling}
\end{equation}
under the common support assumption (the support of $\pi_{\theta}$ support is included in the support of $\pi_{\theta_0}$). The goal in CRM is to find a model $\theta \in \Theta$  with minimal risk by minimizing
\begin{equation}
    \hat{\theta} \in \argmin_{\theta \in \Theta}  \cL_0(\theta), 
        \label{eq:learning_objective_crm}
\end{equation}
where $\cL_0(\theta) = \hat{L}_0(\theta) + \lambda \sqrt{\frac{\hat{V}_0(\theta)}{n}}$ uses the sample variance penalization principle \citep{maurer2009empirical} on samples from $\theta_0$ with counterfactual estimates of the expected risk $\hat L_0$, an empirical variance $\hat V_0$ and $\lambda>0$. Specifically, in the (CRM) framework, multiple estimators are derived from the IPS method~\citep[][]{thompson1952} that uses the following clipped importance sampling estimator of the risk of a model $\theta$ by \citet{bottou2012, swaminathan2012}:
\begin{equation}
   \hat{L}^{\text{IS}}_0(\theta) = \dfrac{1}{n} \sum_{i=1}^{n}  y_{i} \min \left( \dfrac{\pi_\theta(a_{i} |x_{i})}{\pi_{i}}, \alpha \right) , 
       \label{eq:ips}
\end{equation}
where $\alpha$ is a clipping parameter. Writing $\chi_i(\theta)~=~y_{i}~\min(\frac{\pi_\theta(a_{i} |x_{i})}{\pi_{0,i}}, \alpha)$ and $\bar \chi(\theta) = \frac{1}{n} \sum_{i=1}^{n} \chi_i(\theta)$ the empirical variance estimator becomes:
\begin{equation}
   \hat{V}^{\text{IS}}_0(\theta) = \frac{1}{n-1} \sum_{i=1}^{n} (\chi_i(\theta) - \bar \chi(\theta))^2.
       \label{eq:var_ips}
\end{equation}
Other estimators aim at controlling the variance of the estimator with self-normalized estimators \citep{swaminathan2015} or with direct methods \citep{dudik2011, dudik2014} in doubly robust estimators. Even so, the performance of counterfactual learning is harmed when the logging policy under-explores the action space \citep{mcbook}. Likewise, counterfactual estimates obtained from a first round of randomized data collection may not suffice \citep{bottou2012} to select a model $\hat{\theta}$. In those cases, it could be natural to consider collecting additional samples. While it is possible to use the same logging model $\theta_0$ to do so, we will present a framework for designing an improved sequential data collection strategy, following the intuition of sequential designs of \citet{bottou2012}. 

\subsection{Sequential Designs}

In this section we present a design of data collections that sequentially learn a policy from logged data in order to deploy it and learn from the newly collected data. Specifically, we assume that at a round $m \in \lbrace 1, \dots M \rbrace$, a model $\theta_m \in \Theta$ is deployed and a set $s_m$ of $n_m$ observations $s_m~=~(x_{m, i}, a_{m, i}, y_{m,i}, \pi_{m,i})_{i = 1, \ldots, n_m}$ is collected thereof, with propensities $\pi_{m,i}~=~\pi_{\theta_m}(a_{m,i} |x_{m,i})$ to learn a new model $\theta_{m+1}$ and reiterate. In this work, we assume that the loss $y$ is bounded in $[-1, 0]$ as in \citep{swaminathan2012} (note however that this assumption could be relaxed to bounded losses) and follows a fixed distribution $\cP_{\cY}$. Next, we will introduce useful definitions. 

\begin{definition}[Excess Risk and Expected Regret]
Given an optimal model $\theta^*~\in~\argmin_{\theta \in \Theta}~L(\theta)$, we write for each rollout $m$ the excess risk:
\begin{equation}
    \Delta_m = L(\theta_m) - L(\theta^*),
\end{equation}
and define the expected regret as:
\begin{equation}
    R_n = \sum_{m=0}^M \Delta_m n_{m+1}.
\end{equation} 
\end{definition}
The objective is now to find a sequence of models $\lbrace \theta_m \rbrace_{m=1 \dots M}$ that have an excess risk and an expected regret $R_n$ that improve upon CRM guarantees. To do so, we define a sequence of minimization problems for $m \in \lbrace 1, \dots M \rbrace$:
\begin{equation}
    \hat{\theta}_{m+1} \in \argmin_{\theta \in \Theta}  \cL_m(\theta), 
        \label{eq:learning_objective_scrm}
\end{equation}
where $\cL_m$ is an objective function that we define in Section~\ref{sec:learning_strategy}. Note that in the setting we consider, samples are i.i.d inside a rollout $m$ but dependencies exist between different sets of observations. From a causal inference perspective \citep{PetersJanzingSchoelkopf17}, this does not incur an additional bias because of the successive conditioning on past observations. We provide detailed explanations in Appendix \ref{appendix:ignorability-assumption} on this matter. Note also that the main intuition and motivation of our work is to shed light on how learning intermediate models $\theta_m$ to adaptively collect data can improve upon sampling from the same logging system by using the same total sample size $n=\sum_{i=0}^m n_m$. To illustrate the learning benefits of SCRM we now provide a simple example.

\begin{figure*}[ht]
    \centering
\begin{subfigure}
  \centering
    \includegraphics[width=0.3\linewidth]{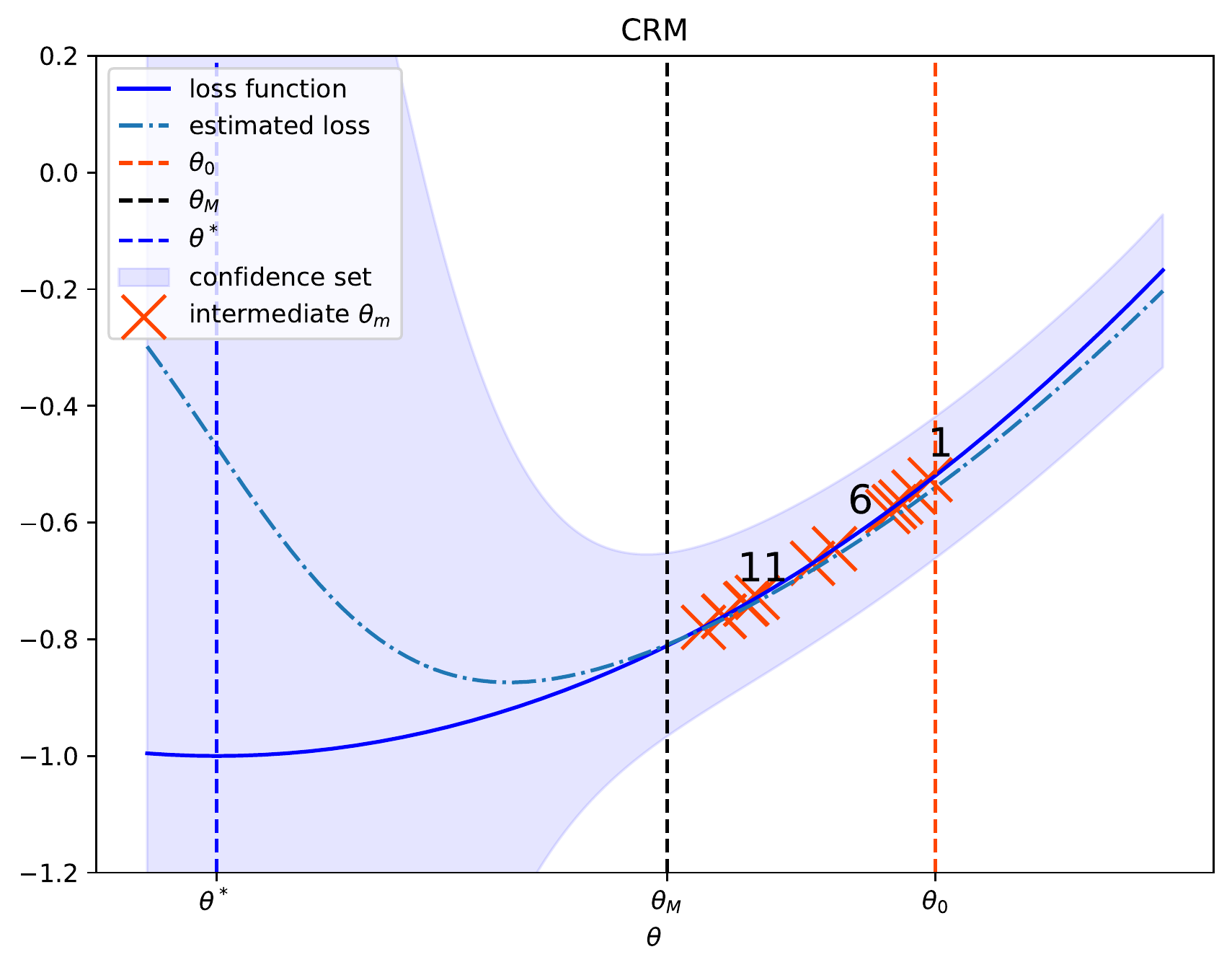}
  \label{fig:sub1}
\end{subfigure}%
\begin{subfigure}
  \centering
    \includegraphics[width=0.3\linewidth]{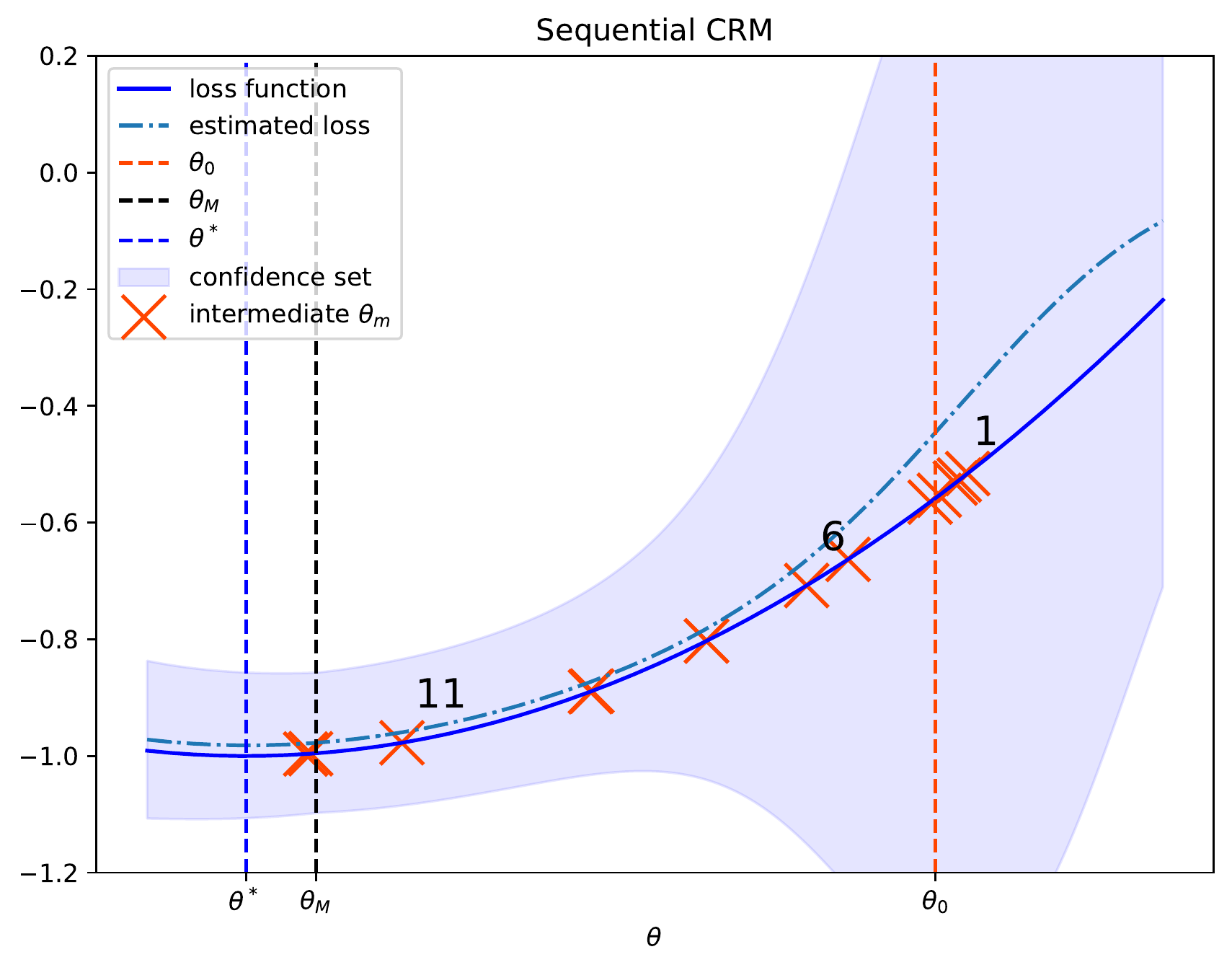}
  \label{fig:sub2}
\end{subfigure}
\begin{subfigure}
  \centering
    \includegraphics[width=0.3\linewidth]{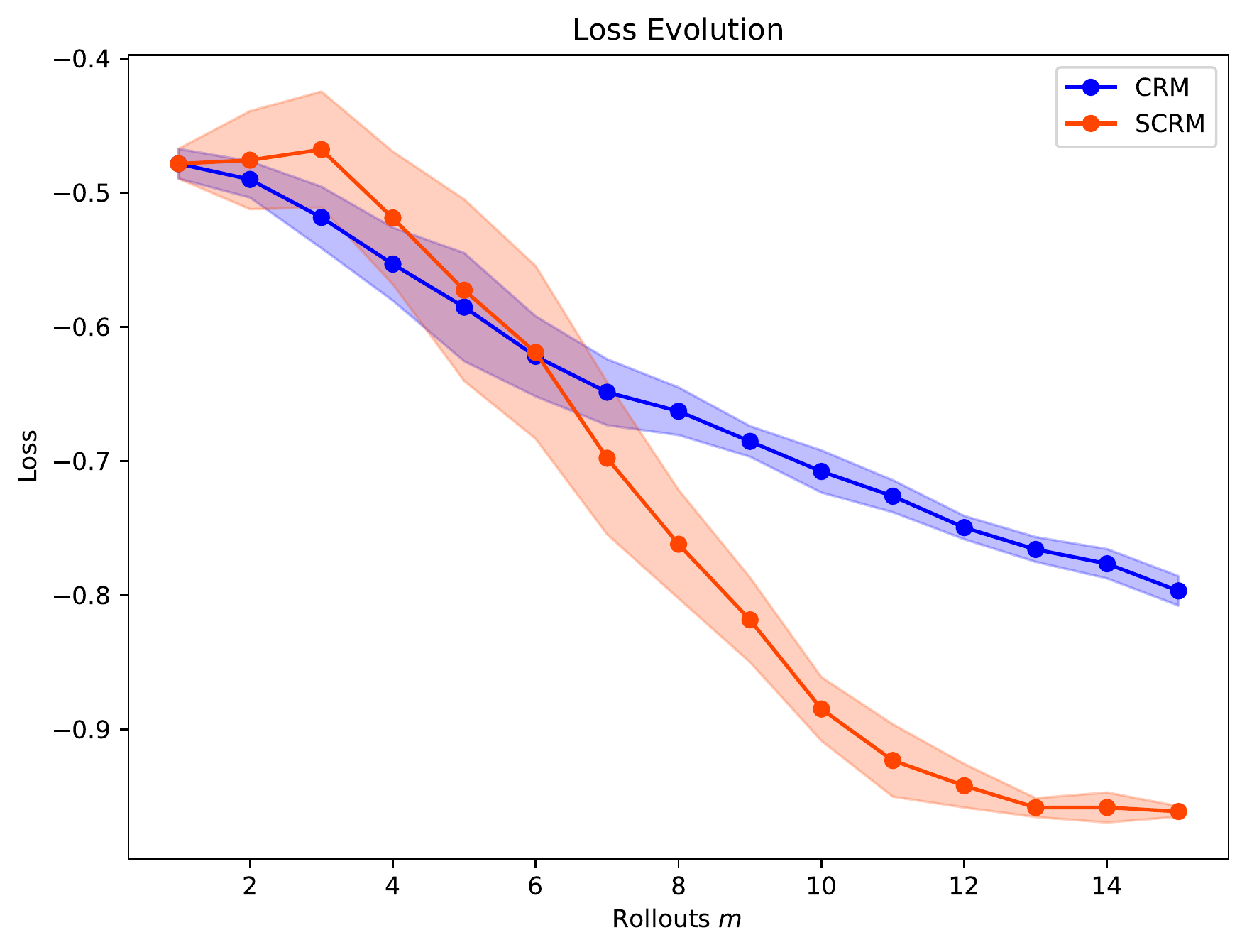}
  \label{fig:sub3}
\end{subfigure}
\caption{Comparison of CRM and SCRM on a simple setting described in Example \ref{example:gaussian_policies}. The models learned through CRM using re-deployments of $\theta_0$ (left) reach $\theta^*$ slower than SCRM (center) that uses intermediate deployments $\theta_1, \dots, \theta_M$ indicated with 'x' markers and rollout numbers. The comparison of the evolution of averaged losses (right) over 10 random runs also shows SCRM converges faster. Here $\theta^*=1$, $\sigma=0.3$ and we take $M=15$ total rollouts with batches $m$ of size $n_m = 100 \times 2^m$. The parameter $\lambda$ is set to its theoretical value.}
\label{fig:gaussian_example}
\end{figure*}

\begin{exmp}[Gaussian policies with quadratic loss]
\label{example:gaussian_policies}
Let us consider Gaussian parametrized policies $\pi_\theta = \mathcal{N}(\theta, \sigma^2)$ and a loss $l_t(a)=(a-y_t)^2-1$ where $y_t \sim \mathcal{N}(\theta^*, \sigma^2)$. We illustrate in Figure \ref{fig:gaussian_example} the evolution of the losses of learned models $\theta_m$ through 15 rollouts with either 
i) Batch CRM learning on aggregation of data, being generated by the unique initial logging policy $\theta_0$ or 
ii) Sequential CRM learning with models $\theta_0, \dots, \theta_{m-1}$ deployed adaptively, with data being generated by the last learned model $\theta_{m-1}$ for the batch $m$. We see that the models learned with SCRM take larger optimization steps than the ones with CRM.  


    
\end{exmp}

We summarize our (SCRM) framework in Algorithm \ref{alg:scrm} with the different blocks exposed previously. We provide an additional graphical illustration of SCRM compared to CRM in Appendix~\ref{appendix:ignorability-assumption}. In the next section we will define counterfactual estimators from the observations $s_m$ at each round and define a learning strategy $\cL_m$. 

\begin{algorithm}[t]
\SetAlgoLined
\KwIn{Logged observations $(x_{0,i}, a_{0,i}, y_{0,i}, \pi_{0,i})_{i = 1, \ldots, n_0}$, parameter $\lambda>0$}

 \For{$m=1$ to $M$}{~
  Build $\cL_m$ from observations  $s_m$ using Eq. \eqref{eq:conservative_learning} \\
 Learn $\theta_{m+1}$ using Eq. \eqref{eq:learning_objective_scrm} \\
 Deploy the model $\theta_{m+1}$ and collect observations $s_{m+1}~=~(x_{m+1,i}, a_{m+1,i}, y_{m+1,i}, \pi_{m+1,i})_{i = 1, \ldots, n_{m+1}}$
 }
 \caption{Sequential Counterfactual Risk Minimization}
 \label{alg:scrm}
\end{algorithm}


\section{Variance-Dependent Convergence Guarantees}

In this part we aim at providing convergence guarantees of counterfactual learning. We show how we can obtain a dependency of the excess risk on the variance of importance weights between the logging model and the optimal model.

\subsection{Implicit exploration and controlled variance}

We first introduce a new counterfactual estimator. For this, we will require a common support assumption as in importance sampling methods \citep{mcbook}. We will assume that the policies $\pi_\theta$ for $\theta \in \Theta$ have all the same support. We then consider the following estimator of the risk of a model~$\theta$:
\begin{equation}
   \hat{L}^{\text{IPS-IX}}_m(\theta) = \dfrac{1}{n_m} \sum_{i=1}^{n_{m}}   \frac{\pi_{\theta, i}}{\pi_{m,i} + \alpha \pi_{\theta, i}} y_{m,i}, 
       \label{eq:ips-ix}
\end{equation}
where $\pi_{\theta, i}=\pi_\theta(a_{m,i} |x_{m,i})$ and $\alpha$ is like a clipping parameter which ensures that the modified propensities $\pi_{m,i} + \alpha \pi_\theta(a_{m,i} |x_{m,i})$ are lower bounded. Noting $\zeta_i(\theta)~= ~\big(\frac{\pi_{\theta, i}}{\pi_{m,i} + \alpha \pi_{\theta, i}} - 1 \big) y_{m,i}$, $\bar \zeta(\theta)~=~\frac{1}{n_m} \sum_{i=1}^{n_m} \zeta_i(\theta)$ we can write the empirical variance estimator as:
\begin{equation}
   \hat{V}^{\text{IPS-IX}}_m(\theta) = \frac{1}{n_m-1} \sum_{i=1}^{n_{m}} (\zeta_i(\theta) - \bar \zeta(\theta))^2.
       \label{eq:var_ips-ix}
\end{equation}
%


%

Here, the empirical variance uses a control variate since it uses the expression of $\zeta_i(\theta)$ above instead of $y_{m,i} \frac{\pi_{\theta, i}}{\pi_{m,i} + \alpha \pi_{\theta, i}}$. This allows to improve the dependency on the variance in the excess risk provided in Proposition \ref{prop:excess_risk}. Note also that our estimator resembles the implicit exploration estimator in the EXP3-IX algorithm \citep{lattimore2019}, as our motivation is to improve the control of the variance.

\subsection{Learning strategy}
\label{sec:learning_strategy}

Next, we aim in this part to provide a learning objective strategy $\mathcal{L}_m$, as referred to in Eq. \eqref{eq:learning_objective_scrm}. Our approach, like the (CRM) framework, uses the sample variance penalization principle \citep{maurer2009empirical} to learn models that have low expected risk with high probability. To do so, we first provide an assumption to be used in our generalization error bound. 

\begin{assum}[Bounded importance weights]
For any models $\theta, \theta' \in \Theta$ and any  $(x, a) \in \cX \times \cA$, we assume $ {\pi_{\theta}(a| x)}/{\pi_{\theta'}(a| x)} \leq W$, for some $W >0$.
\label{assum:bounded_ipw}
\end{assum}

This assumption has been made in previous works \citep{Kallus2018PolicyEA, zenati_counterfactual} and is reasonable when we consider a bounded parameter space $\Theta$. Next, we state an error bound for our estimator.

\begin{proposition}[Generalization Error Bound]
{\label{prop:generalization_bound}}
Let $\hat{L}^{\text{IPS-IX}}_m$ and $\hat{V}^{\text{IPS-IX}}_m$ be the empirical estimators defined respectively in Eq. \eqref{eq:ips-ix} and Eq. \eqref{eq:var_ips-ix}.  Let $\theta \in \Theta$, $\delta \in (0,1)$, and $n_m \geq 2$. Then, under Ass.~\ref{assum:bounded_ipw}, for $\lambda_m=\sqrt{18(C_m(\Theta)+ \log(2/\delta))}$, with probability at least $1 - \delta$:
\begin{equation*}
      L(\theta) \leq \hat{L}^{\text{IPS-IX}}_m(\theta) + \lambda_m \sqrt{\frac{\hat{V}^{\text{IPS-IX}}_m(\theta)}{n_m}} + \frac{2\lambda_m^2W}{n_m} + \delta_m,
\end{equation*}
where $C_{m}(\Theta)$ is a metric entropy complexity measure defined in App.~\ref{appendix:analysis_definition} and $\delta_m = \sqrt{\log(2/\delta)/({2n_m})}$.

\end{proposition}

This Proposition is proved in Appendix \ref{appendix:analysis-upper-bound} and essentially uses empirical bounds \citep{maurer2009empirical}. By minimizing the latter high-probability upper bound, we can find models $\theta$ with guarantees of minimizing the expected risk. Therefore, at each round, we minimize the following loss: 
\begin{equation}
    \mathcal{L}_m(\theta) = \hat{L}^{\text{IPS-IX}}_m(\theta) + \lambda_m \sqrt{\frac{\hat{V}^{\text{IPS-IX}}_m(\theta)}{n_m}},
\label{eq:conservative_learning}
\end{equation}
where $\lambda_m>0$ is a positive parameter. Unlike deterministic decision rules used for example in UCB-based algorithms \citep{lattimore2019}, the exploration is naturally guaranteed by the stochasticity of the policies we use.   



\subsection{Excess risk upper bound}

Eventually, we establish an upper bound on the excess risk of the IPS-IX estimator for counterfactual risk minimization using the learning strategy that we just defined. For this, we require an assumption on the complexity measure.

\begin{assum}
\label{assumption:logarithmic_complexity}
    We assume that the set $\Theta$ is compact and that there exists $d > 0$ such that
    $
        C_m (\Theta) \leq  d \log(n_m).
    $
\end{assum}

This assumption states that the complexity grows logarithmically with the sample size. It holds for parametric policies so long as the propensities are lower bounded, which is verified using our estimator. We now state our variance-dependent excess risk bound.

\begin{proposition}[Excess Risk Bound]
Let $n_m\geq 1$ and $\theta_m \in \Theta$. Let $s_m$ be a set of $n_m$ samples collected with policy $\pi_{\theta_m}$. Then, under Assumptions~\ref{assum:bounded_ipw} and~\ref{assumption:logarithmic_complexity}, a minimizer $\theta_{m+1}$ of  Eq.~\eqref{eq:conservative_learning} on the samples $s_m$ satisfies the excess risk upper-bound:  w.p. $1-\delta$
\begin{align*}
    & \Delta_{m+1}  =~L(\theta_{m+1}) - L(\theta^*) \\
    & \lesssim   \sqrt{\nu_m^2 \frac{d\log n_m \! - \! \log \delta}{n_m}}  + \frac{W^2 + W (d \log n_m\! - \!\log \delta)}{n_m},
\end{align*}
where $\nu_m^2 = \Var_{x, \theta_{m}}\left(\dfrac{\pi_{\theta^*}(a|x)}{\pi_{\theta_{m}}(a|x)}\right)$.
\label{prop:excess_risk}
\end{proposition}
The proof is postponed to  Appendix \ref{appendix:analysis-upper-bound}. The modified propensities in IPS-IX as well as the control variate used in the variance estimator allow us to improve the dependency in $\nu_m^2$, compared to $\nu_m^2 + 1$ obtained in previous work~\citep{zenati_counterfactual}. This turns out to be a crucial point to use these error bounds sequentially as in acceleration methods since $\nu_m \to 0$ if $\theta_m \to \theta^*$, as explained in the next section.

\section{SCRM Analysis}
\label{sec:regret_analysis}

In this section we provide the main theoretical result of this work on the excess risk and regret analysis of SCRM. We start by stating an assumption that is common in acceleration methods \citep{daspremont21} with restart strategies \citep{becker2011,nesterov2012GradientMF} that we will require to achieve the benefits of sequential designs.

\begin{assum}[H\"olderian Error Bound]
We assume that there exist $\gamma>0$ and $\beta >0$ such that for any $\theta \in \Theta$, there exists $\theta^* \in \argmin_{\theta \in \Theta} L(\theta)$ such that
\begin{equation*}
   \gamma \Var_{x, \theta} \left( \dfrac{\pi_{\theta^*}(x|a)}{\pi_\theta(x|a)} \right) \leq \left( L(\theta) - L(\theta^*) \right)^\beta \,.
\end{equation*}
\label{assum:holderian_error_bound}
\end{assum}
Typically, in acceleration methods, H\"olderian error bounds \citep{bolte2007} are of the form:
\begin{equation*}
    \gamma d(\theta, S_\Theta^*) \leq \left( L(\theta) - L(\theta^*) \right)^\beta
\end{equation*}
for some $\gamma, \beta > 0$ and where $d(\theta, S_\Theta^*)$ is some distance to the optimal set ($S_\Theta^* = \argmin_{\theta \in \Theta} L(\theta)$). This  bound is akin to a local version of strong convexity ($\beta=1$) or a bounded parameter space ($\beta=0$) if $d$ is the Euclidean distance. When $\beta \in [0,1]$, this has also been referred to as the Łojasiewicz assumption introduced in \citep{lojasiewicz1963, lojasiewicz1993}. Notably, it has been used in online learning \citep{gaillard_2018} to obtain fast rates with restart strategies. 
%
%
This assumption holds for instance for Example~\ref{example:gaussian_policies} with $\beta = 1$ (see App~\ref{app:gaussian_example}). 
We also discuss this assumption for distributions in the exponential family in Appendix \ref{appendix:exponential_families} notably for distributions that have been used practice \citep{swaminathan2015, Kallus2018PolicyEA, zenati_counterfactual}. 
Next we state our main result that is the acceleration of the excess risk convergence rate and the regret upper bound of SCRM.

\begin{proposition}
 Let $n_0, n \geq 2$ and $\theta^* \in \argmin_{\theta} L(\theta)$. Let $n_m = n_02^m$ for $m=0,\dots,M  = \big\lfloor \log_2 ( 1+ \frac{n}{n_0}) \big\rfloor$. Then, under Assumptions~\ref{assum:bounded_ipw},~\ref{assumption:logarithmic_complexity} and~\ref{assum:holderian_error_bound} with $\beta >0$,  the SCRM procedure (Alg. \ref{alg:scrm}) satisfies the excess risk upper-bound
   \begin{equation*}
      \Delta_M = L(\theta_M) - L(\theta^*) \leq O\Big( n^{-\frac{1}{2-\beta}}\log n\Big) .
   \end{equation*}
   Moreover, the expected regret is bounded as follows:
   \begin{equation*}
    R_n =  \sum_{m=0}^M  \Delta_m n_{m+1}   \leq O\Big( n^{\frac{1-\beta}{2-\beta}}\log(n)^2\Big) .
\end{equation*}
\label{prop:regret_bound}
\end{proposition}
The proof of our result is detailed in Appendix \ref{appendix:regret_proof}.

\paragraph{Discussion} This result illustrates that an excess risk of order $O\big(\frac{\log(n)}{n}\big)$ may be obtained when $\beta=1$ (which is implied by a local version of strong convexity assumption in acceleration methods). When $\beta =0$, which merely accounts that the variance of importance weights are bounded, we simply recover the original rate of CRM of order $O(\log(n)/\sqrt{n})$. The SCRM procedures thus improves the excess risk rate whenever $\beta>0$. It is worth to emphasize that the knowledge of $\beta$ is not needed by Alg.~\ref{alg:scrm}. We also note that our assumption seems related to the Bernstein condition \citep[see Def 2.6]{bartlett2006}, and \citep[see Def 5.1]{vanerven15a} that bounds a variance term by an excess risk term to the power. In empirical risk minimization, this implies the same excess risk rate and regret rate \citep{vanerven16}, which are exactly the same rates as ours (up to logs).

\section{Empirical Evaluation}

In this section we perform numerical experiments to validate our method in practical settings. We present the experimental setup as well as experiments comparing SCRM to related approaches and internal details of the method.

\subsection{Experimental setup}

As our method is able to handle both discrete and continuous actions we  experiment in both settings.
We now provide a brief description of the setups, with extensive details available in Appendix \ref{appendix:experiment_settings_details}. \footnote{All the code to reproduce the empirical results is available at: \url{https://github.com/criteo-research/sequential-conterfactual-risk-minimization}}

\paragraph{Continuous actions} We perform evaluation on synthetic problems pertaining to 
personalized pricing problems from \citep{demirer2019semi} (\benchpricing) and advertising 
from \citep{zenati_counterfactual} (\benchpotential).
We consider Gaussian policies $\pi_\theta( \cdot | x) = \cN(\mu_\theta(x), \sigma^2) $ with linear contextual parametrization $\mu_\theta(x)=\theta^\top x$ and fixed variance $\sigma^2$ that corresponds to the exploration budget allowed in the original randomized experiment. The features are up to 10 dimensions and the actions are one-dimensional. We keep the original logging baselines from the settings and compare results to a skyline supervised model trained on the whole training data with full information. 

\paragraph{Discrete actions} We adapt the setup of \citep{swaminathan2012} that transforms a multilabel classification task into a contextual bandit problem with discrete, combinatorial action space. We keep the original modeling (akin to CRF) with categorical policies $\pi_\theta( a | x) \propto \exp(\theta^\top (x \bigotimes a))$. The baseline (resp. skyline) is a supervised, full information model with identical parameter space than CRM methods trained on 5\% (resp. 100\%) of the training data.
We consider the class of probabilistic policies that satisfy Assumption \ref{assum:holderian_error_bound} by predicting actions in an Epsilon Greedy fashion \cite{sutton1998}): $\pi^{\epsilon}_\theta(a,x) = (1-\epsilon) \pi_\theta(a,x) + \epsilon/|\mathcal{A}|$ where $\epsilon=.1$.
Real-world datasets include \benchscene, \benchyeast\ and \benchtmc\ with feature space up to 30,438 dimensions and action space up to $2^{22}$.
To account for this combinatorial action space we allow a model $\theta_m$ to be learned using data from all past rollouts $\{s_l\}_{l < m}$ for better sample efficiency and therefore adjust variance estimation in Appendix \ref{appendix:multi_ips} to take into account sequential dependencies. 


\subsection{SCRM compared to CRM and related methods}

We first compare SCRM to CRM and existing methods in the literature.
\begin{figure*}[t]
    \centering
    
    \begin{minipage}{0.24\textwidth}
        \includegraphics[width=0.9\linewidth]{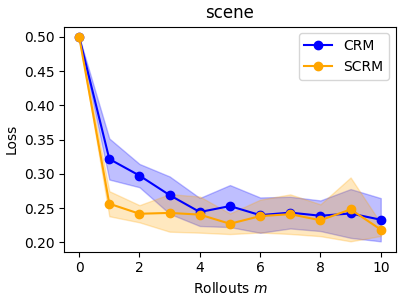}
    \end{minipage}\hfill
    \begin{minipage}{0.24\textwidth}
        \includegraphics[width=0.9\linewidth]{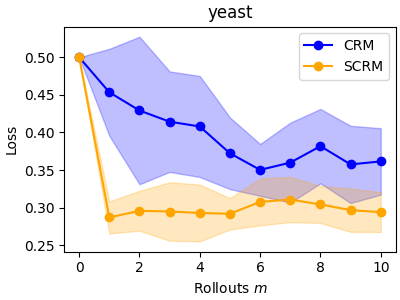}
    \end{minipage}
    \begin{minipage}{0.24\textwidth}
        \includegraphics[width=0.9\linewidth]{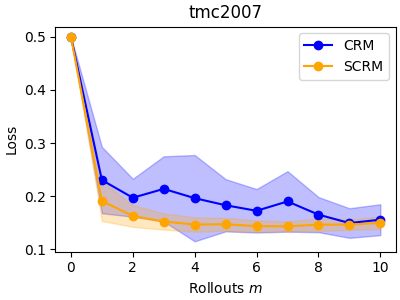}
    \end{minipage}\hfill
    \begin{minipage}{0.24\textwidth}
        \includegraphics[width=0.9\linewidth]{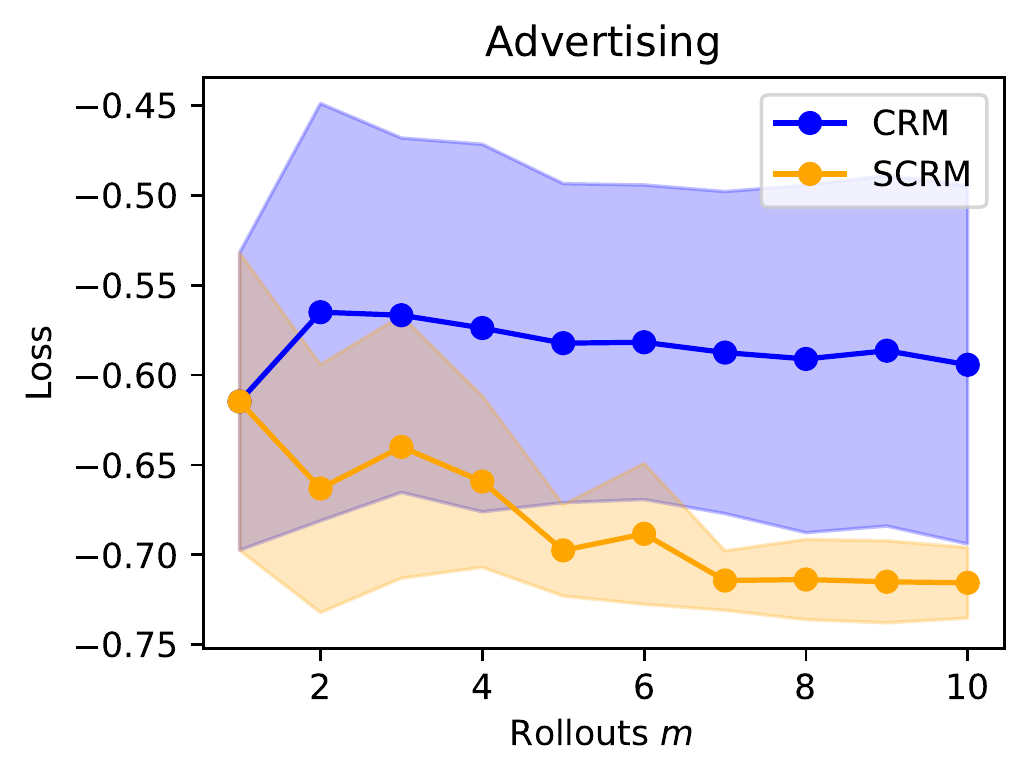}
    \end{minipage}
    \caption{Test loss as a function of sample size on \benchscene, \benchyeast, \benchtmc, \benchpotential,\ (from left to right). SCRM (in orange) converges faster and with less variance than CRM (in blue).}
    \label{fig:scrm_vs_crm}
\end{figure*}
\paragraph{Comparison between SCRM and CRM} First, we provide insights on the performance that SCRM can achieve compared to classical CRM with increasing sample sizes. The key difference between CRM/SCRM is that for each sample size $n_m$ CRM learns from samples generated by the logging model $s^{CRM}_m \leftarrow \theta_0$ (see Alg. \ref{alg:crm}) whilst SCRM learns from samples generated by a series of optimized models $s^{SCRM}_m \leftarrow \theta_m$ (see Alg. \ref{alg:scrm}). For each sample size we select a posteriori the best $\lambda$ for both methods based on test set loss value. We report in Figure \ref{fig:scrm_vs_crm} over $M=10$ rollouts the mean test loss depending on sample size up to $2^{10}$, with standard deviation estimated over 10 random runs.
We observe that SCRM converges very fast, often within the first rollouts. Conversely, CRM needs more samples and the variance is higher. We conclude that there is a striking benefit to use a sequential design in order to achieve near optimal loss with much fewer samples and better confidence compared to CRM.
Complementary results on other datasets are available in Appendix \ref{appendix:scrm_vs_crm}.

\begin{table}[t]
    \centering
    \begin{tabular}{l|l|l|l}
         Percentage $p$        & $0.7$ & $0.8$ & $0.9$ \\ 
        \hline
         CRM     &  $100\times 2^{10}$ & $100\times 2^{16}$ & $> 100\times 2^{22}$ \\ \hline
         SCRM (ours)    & $100\times 2^8$  & $100\times 2^9$ & $100\times 2^{11}$ \\ \hline
    \end{tabular}
    \caption{Needed sample size to achieve test loss $L(\theta)~\leq~p*L(\theta^*)$ on the setting in Example \ref{example:gaussian_policies} over the average of 10 random runs. SCRM needs way less data to converge to near optimal solution. $\lambda$ is set to its theoretical value.}
    \label{tab:sample_size_needed}
\end{table}

Moreover, to further illustrate this benefit of efficient learning we also report in Table \ref{tab:sample_size_needed} the sample size needed to attain near optimal performance when $\theta^*$ is known as in Example \ref{example:gaussian_policies}, where we also observe that SCRM reaches optimal performances faster than CRM. This corroborates the benefits of improved excess risk rates for SCRM. 

\begin{table}[t]
    \centering
    \resizebox{\linewidth}{!}{  
    \begin{tabular}{l|r|r|r|r}
                 & \benchpricing & \benchpotential & \benchyeast & \benchtmc \\ 
        \hline
         $\lambda'$        & $-5.353 \pm .178$ & $-.716 \pm .020$ & $.294 \pm .026$ & $.146 \pm .012$ \\ \hline
         $\hat{\lambda}$    & $-5.575 \pm .036$ & $-.726 \pm .001$ & $.299 \pm .039$ & $.164 \pm .021$ \\ \hline
    \end{tabular}
    }
    \caption{Test loss after 10 rollouts when choosing $\lambda$ by a posteriori selection ($\lambda'$) or with proposed heuristic ($\hat{\lambda}$). Our heuristic is competitive with the a posteriori selection of a fixed $\lambda'$.}
    \label{tab:crossval_lambda}
\end{table}

\paragraph{Hyper-parameter selection for SCRM} In our experiments, hyperparameter selection consists in choosing a value for $\lambda$. We describe a simple heuristic and evaluate its performance on different datasets. We propose to select $\hat{\lambda}_m$ by estimating the non-penalized CRM loss (eq. \ref{eq:learning_objective_crm}) using offline cross-validation on past data $s_{t<m}$.
We report in Table \ref{tab:crossval_lambda} the test loss obtained when choosing a fixed $\lambda$ a posteriori ($\lambda'$) or with this heuristic ($\hat{\lambda}$). We observe that loss confidence intervals for both methods intersect for all discrete datasets, except on \benchtmc \ where the degradation shows only at the 3rd digit. On continuous datasets, the heuristic actually improves upon the fixed a posteriori selection. We conclude that this heuristic is usable in practice.


\begin{table*}[t!]
\centering
\begin{tabular}{l|r|r|r|r|r}
            & \benchpricing & \benchpotential & \benchscene & \benchyeast & \benchtmc \\
 $n/|\cA|/dim(\cX)|$ & $10^5/\infty/10$ & $10^5/\infty/2$ & $2.10^3/2^6/295$ & $2.10^3/2^{14}/104$ & $3.10^4/2^{22}/3.10^4$\\ \hline \hline
Baseline    & $-3.414 \pm .162$ & $-.431 \pm .120$  & $.353 \pm .009$ & $.478 \pm .014$ & $.511 \pm .003$ \\ \hline
\hline
SBPE        &   DNF           &    DNF         &    $\textbf{.179} \pm .001$             &      $.302 \pm .003$           & DNF         \\ \hline
BKUCB       &   DNF           &   DNF          &      $.236 \pm .014$           &      $.303 \pm .004$           & DNF         \\ \hline
\hline
TRPO        & $ \textbf{-5.750} \pm .020$ & $-.670 \pm .030$ & $.376 \pm .001$ & $.434 \pm .001$ & $.396 \pm .001$  \\ \hline
PPO         & $-5.274 \pm .200$ & $-.637 \pm .015$ & $.206 \pm .001$ & $.463 \pm .001$ & $.263 \pm .001$  \\ \hline
\hline
CRM         & $-5.325 \pm .068$ & $-.594 \pm .100$ & $.233 \pm .031$ & $.362 \pm .044$ & $.158 \pm .034$  \\ \hline
SCRM (ours) & $ -5.575 \pm .036$ & $\textbf{-.726} \pm .020$ & $.219 \pm .009$ & $\textbf{.294} \pm .026$ & $\textbf{.146} \pm .012$  \\ \hline 
\hline
Skyline     & $-5.830 \pm .020$ & $-.739 \pm .002$ & $.179 \pm .002$ & $.312 \pm .003$ & $.142 \pm .001$  \\ \hline
\end{tabular}
\caption{Test loss $\pm$ stddev of different methods after 10 rollouts. SCRM achieves optimal or near optimal performance in all datasets. Batch bandit methods did not finish (DNF) on large scale settings, and RL methods perform overall poorly on discrete settings with large action space.}
\end{table*}

\paragraph{Comparison with other methods} In this paragraph we compare our SCRM to related methods to explore practical implications of existing methods in our setting. We first consider batch bandits methods and implement the stochastic sequential batch pure exploitation (SBPE) algorithm in \citep{han2020sequential} and a batch version of kernel UCB \citep{valko_2013} algorithm (BKUCB) with an optimized library (see implementations details in Appendix \ref{appendix:experiment_implementation_details}). We also experiment with off-policy RL methods PPO \citep{PPOref} and TRPO \citep{TRPOref} from the StableBaselines library \citep{stable-baselines3} (see Appendix \ref{appendix:experiment_implementation_details}). Indeed, such methods model more general state transitions based on past actions, but they could be used in our setting. To fairly compare all methods (in particular those for which no heuristic existing for hyper-parameter selection) we report the mean and standard deviation over 10 random runs of the best test loss a posteriori over hyperparameter grids of the same size. First, we observe that SCRM beats CRM on all datasets, illustrating the benefit of the sequential design. Second, on discrete tasks (where we the combinatorial action space is large) we observe that SCRM achieves nearly the best test loss in all tasks, while RL methods have difficulties maintaining good performances. Third, batch bandits algorithms can achieve good performances in practice because of their deterministic decision rules. However, they involve an $O(n^3)$ matrix inversion and therefore did not finish (DNF) in 24h (per single run) on a 46 CPU / 500G RAM machine in most of our settings with large sample size $n$, which make them unpractical for large scale experiments. 
We conclude that SCRM is an effective learning paradigm and that it scales successfully on a variety of settings.

\subsection{Details on SCRM}
Next, we provide additional empirical evaluations of details of our method.
\label{sec:details_scrm}

\begin{figure}[h]
    \centering
    \includegraphics[width=\columnwidth]{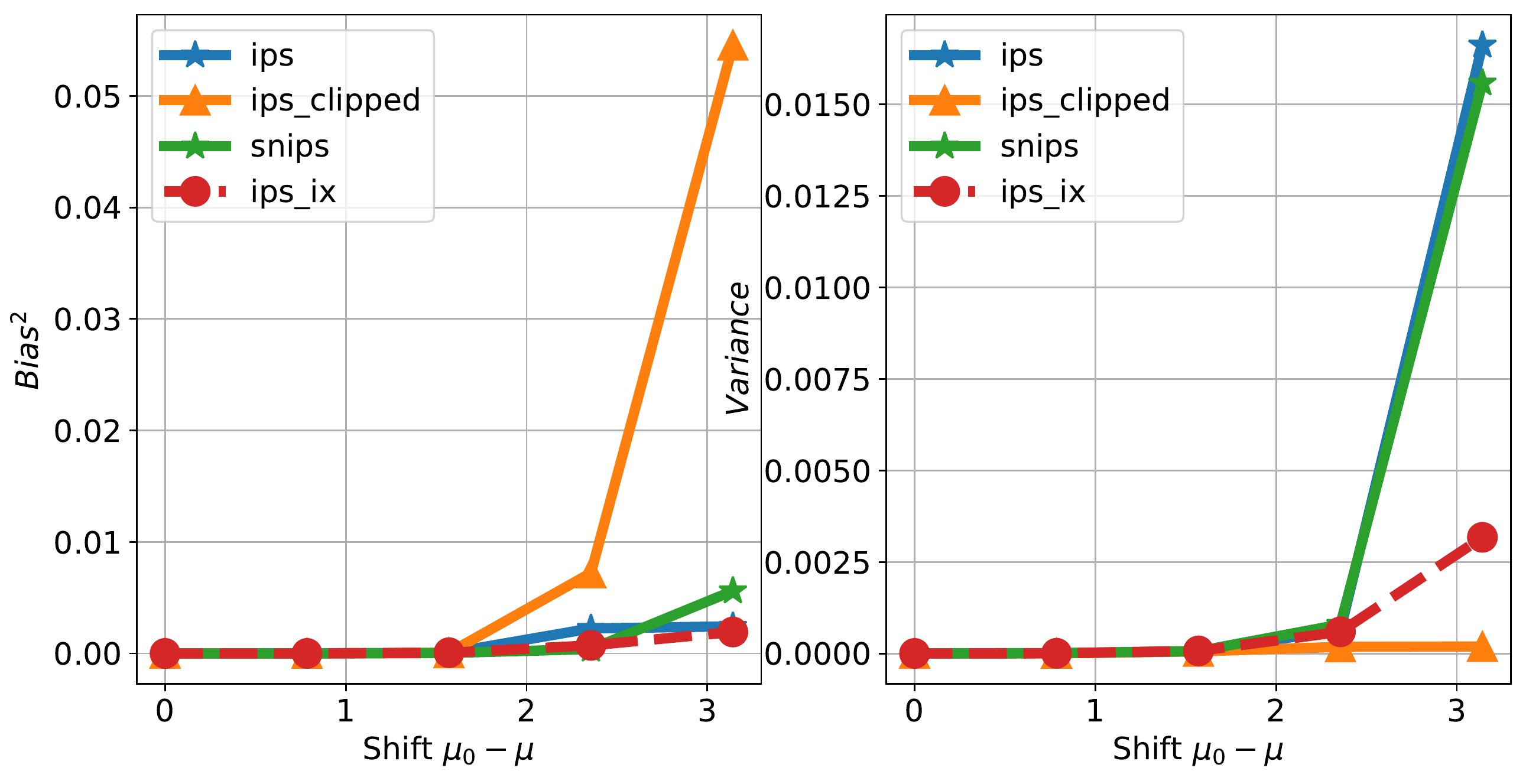}       
    \caption{Comparison of counterfactual estimators on policy evaluation. Bias (left), Variance (right). IPS-IX shows a low bias and compares favorably to IPS and SNIPS in terms of variance.}
    \label{fig:compare_estimators}
\end{figure}

\paragraph{Evaluation of IPS-IX} To understand the bias-variance trade-off that IPS-IX can achieve in practice compared to other counterfactual estimators we consider a policy evaluation experiment. The task we consider uses sinusoidal losses $y(a)=\cos(a)$ and evaluated policies are shifted Gaussians $\{\pi_i = \mathcal{N}(i*\pi/4,1)\}_{i=0,4}$, with $\pi_0$ being the logging policy. Evaluated policies with large shifts with $\pi_0$ therefore simulate the setting where the logging policy under-explores the action space. The estimators we consider include IPS, SNIPS \cite{swaminathan2015}, clipped IPS (eq. \ref{eq:ips}) with heuristic from \cite{bottou2012} and IPS-IX (eq. \ref{eq:ips-ix}) with $\alpha=1/n$. All methods therefore use their respective heuristics to set hyperparameters. We report in Figure \ref{fig:compare_estimators} the bias and variance of estimators for each shift $\mu_0 - \mu = i*\pi/4$ for $i=0, \dots, 4$.
We observe that IPS-IX shows an empirical bias comparable to IPS, lower than SNIPS and clipped IPS while maintaining a lower variance. Moreover its variance is only slightly higher than clipped IPS which introduced a large bias. We conclude that besides being a key component of our analysis IPS-IX also controls the variance with a better trade-off in practice. More details are available in Appendix \ref{appendix:evaluation_ips_ix}.

\begin{figure}
    \centering
    \includegraphics[width=0.6\columnwidth]{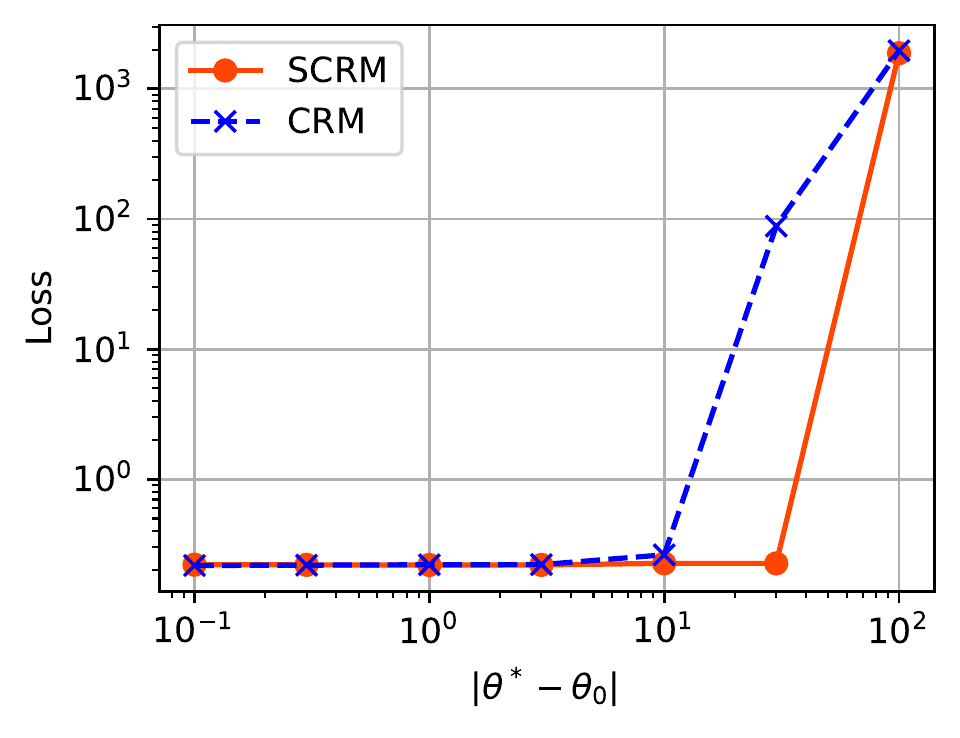}
    \caption{Best final loss when varying $\delta_0 = \Vert \theta^* - \theta_0 \Vert$. SCRM achieves better losses especially for larger $\delta_0$.}    \label{fig:distance_variance_compromise}
\end{figure}

\paragraph{When is SCRM useful}
is a natural question of interest when choosing the method to be used on a given logged bandit feedback problem. Intuitively one can imagine that SCRM will be most useful when the logging policy underexplores the action space, for example when the distance (in parameter space) between the logging and optimal parameters is large. To study this question we proceed to the following experiment on the setup of Example \ref{example:gaussian_policies} with Gaussian distributions $\cN(\theta, \sigma)$ and fixed loss variance $\sigma^*=\Var_y(y)$.
We vary the distance $\delta_0 = \Vert \theta^* - \theta_0 \Vert$ between the optimal model $\theta^*$ and the logging model $\theta_0$. Since the ideal exploration level may be task dependent we choose a posteriori the best $\sigma$ on a grid, for both CRM and SCRM. 
We report in Figure~\ref{fig:distance_variance_compromise} the best final loss for both CRM and SCRM for a range of values of $\delta_0$. 
We observe in particular that SCRM achieves better final losses for larger distances $\delta_0$ than CRM. With the same number of rollouts $M$, SCRM can extend the exploration to further areas while CRM fails for any exploration level in those cases, which advocates for using sequential deployments.

\section{Discussions}

In this work, we have proposed a method to extend the CRM perspective for designing sequential data collection experiments. We have introduced a novel counterfactual estimator to improve variance control in excess risk bounds. Under a weak error bound assumption, we have sequentially applied these excess risk guarantees to achieve faster rates similarly to acceleration methods. Our method also improves upon CRM in practice and is particularly well-suited for this setting compared to existing methods in the literature. It is worth noting that, in order to avoid introducing dependencies in the excess risk bounds we analyzed, the theoretical algorithm we have studied uses geometric sample sizes to discard previous samples. However, using all past samples has been found to be also effective in practice and developing guarantees for this case would be an interesting area for future research. Additionally, similar to online settings that involve an exploration-exploitation tradeoff, investigating the use of optimism in the face of uncertainty (OFUL) principle in SCRM would also be a promising avenue for future work.


\section*{Acknowledgements}

The authors thank Alberto Bietti for the insightful early discussions on this project. The authors also thank the reviewers for their feedback on this paper. This work was supported by ANR 3IA MIAI@Grenoble-Alpes (ANR-19-P3IA0003).


\bibliography{references}
\bibliographystyle{icml2023}

\newpage
\appendix
\onecolumn

\input{appendix}

\end{document}

%% file: related_work.tex

Counterfactual learning from logged feedback \citep{bottou2012} uses only past interactions to learn a policy without interacting with the environment. Counterfactual risk minimization methods  \citep{swaminathan2012, swaminathan2015} propose learning formulations using a variance penalization as in \citep{maurer2009empirical} to find policies with minimal variance. Even so, counterfactual methods remain prone to large variance issues \citep{dudik2014}. These problems may arise when the logging policy under-explores the action space, making it difficult to use importance sampling tehcniques \citep{mcbook} that are key to counterfactual reasoning. While one could collect additional data to counter this problem, our method focuses on sequential deployments \citep[see Section 6.4]{bottou2012} to collect data obtained from adaptive policies to explore the action space. Note also that the original motivation is related but different from the support deficiency problem \citep{thorsten2020defficientsupport} where the support of the logging policy does not cover the support of the optimal policy. 

Another related literature to our framework is batch bandit methods. Originally introduced by \citet{Perchet2015} and then extended by \citet{Gao2019} in the multi-arm setting, batch bandit agent take decisions and only observe feedback in batches. This therefore differs from the classic bandit setting \citep{auer2002, Audibert2007TuningBA} where rewards are observed after each action taken by an agent. Extensions to the contextual case have been proposed by \citet{han2020sequential} and could easily be kernelized \citep{valko_2013}. The sequential counterfactual risk minimization problem is thus closely related to this setting. However, major differences can be noted. First, SCRM does not leverage any problem structure as in stochastic contextual bandits \citep{li2010} by assuming a linear reward function \citep{Chu2011, Goldenshluger2013, han2020sequential} nor uses regression oracles as \citep{foster20a, SimchiLevi2020}. Second, deterministic decision rules taken by bandit agents \citep{lattimore2019} do not allow for counterfactual reasoning or causal inference \citep{PetersJanzingSchoelkopf17}, unlike our framework which performs sequential randomized data collection. Third, unlike gradient based methods used in counterfactual methods with parametric policies, batch bandit methods use zero-order methods to learn from data and necessitate approximations to be scalable \citep{calandriello20a, zenati22a}.

The sequential designs that we use are adaptive data collection experiments, which have been studied by \citet{Bakshy2018AEAD, KasySautmann}. Closely related to our method is policy learning from adaptive data that has been studied by \citet{zhan2021policy} and \citet{Kallus2021} in the online setting. In contrast, we consider a batch setting and our analysis achieve fast rates in more general conditions. \citet{zhan2021policy} use a doubly robust estimator and provide regret guarantees but assume a deterministic lower bound on the propensity score to control the variance. Instead, our novel counterfactual estimator does not require such an assumption. \citet{Kallus2021} propose a novel maximal inequality and derive thereof fast rate regret guarantees under an additional margin condition that can only hold for finite action sets. Our work instead uses a different assumption on the expected risk, which is similar to H\"olderian error bounds in acceleration methods \citep{daspremont21} that are known to be satisfied for a broad class of subanalytic functions \citep{bolte2007}.

In the reinforcement learning literature \citep{sutton1998}, off-policy methods \citep{offpolicy_harutyunyan2016, offpolicy_munos_2016} evaluate and learn a policy using actions sampled from a behavior (logging) policy, which is therefore closely related to our setting. Among methods that have shown to be empirically successful are the PPO \citep{PPOref} and TRPO \citep{TRPOref} algorithms which learn policies using a Kullback-Leibler distributional constraint to ensure robust learning, which can be compared to our learning strategy that improves the logging policy at each round. However reinforcement learning models transitions in the states (contexts) induced by the agent's actions while bandit problems like ours assume that actions do not influence the context distribution. This enables to design algorithms that exploit the problem structure, have theoretical guarantees and can achieve better performance in practice. 

Finally, our method is related to acceleration methods \citep{daspremont21} where current iterates are used as new initial points in the optimization of strongly convex functions \citep{boyd_vandenberghe_2004}. While different schemes use fixed \citep{PowellRestart} or adaptive \citep{NoceWrig06, becker2011, nesterov2012GradientMF, bolte2007, gaillard_2018} strategies, our method differs in that it does not consider the same original setting, does not require the same assumptions nor provides the same guarantees. Eventually, while current models are also used as new starting points, additional data is effectively collected in our setting unlike those previous works that do not assume partial feedbacks as in our case.

%% file: appendix.tex
This appendix is organized as follows: in Appendix \ref{appendix:counterfactual_methods_details}, we provide additional explanations on counterfactual methods related to our approach. In Appendix \ref{appendix:analysis-details}, we detail our analysis of our counterfactual estimator as well as the general SCRM procedure, as given in Alg. \ref{alg:scrm}. Next, in Appendix \ref{appendix:experiment_details} we present all the details of the empirical evaluation and eventually in Appendix \ref{appendix:additional_empirical_results} we provide all additional empirical results that were omitted from the main paper due to space limitation. 

\section{Additional details on counterfactual estimators}
\label{appendix:counterfactual_methods_details}

\subsection{Unconfoundedness in sequential designs}

\label{appendix:ignorability-assumption}

\begin{figure}[htbp]
\begin{center}
\resizebox{0.3\columnwidth}{!}{
\begin{tikzpicture}[auto, thick, node distance=1cm, >=triangle 45]
\tikzstyle{unobserved}=[thick, dashed, fill=gray!0]
\tikzstyle{squared}=[thick, fill=gray!10, rounded corners=5pt]
\tikzstyle{norn}=[thick, fill=gray!0, minimum size=1cm]
\draw[squared] (-2.9,-0.9) rectangle (-1.1,0.9);
\draw[squared, fill=blue!20] (-0.9,-0.9) rectangle (0.9,0.9);
\draw[squared, fill=blue!30] (1.1,-0.9) rectangle (2.9,0.9);
\draw[squared] (3.1,-0.9) rectangle (4.9,0.9);
\draw
	node[norn] at (2,0)[causalvar](A){$A$}
	node[norn] at (4,0)[causalvar](Y){$Y$}
	node[norn] at (-2,0)[causalvar](X){$X$}
    node[squared, minimum size=1cm] at (0,0)[causalvar](theta){$\theta$};
	
\draw (0,-0.7)node[scale=0.7]{model};
\draw (2,-0.7)node[scale=0.7]{treatment};
\draw (-2.,-0.7)node[scale=0.7]{context};
\draw (4,-0.7)node[scale=0.7]{outcome};
	\draw[->](X) --  (theta) node[midway, near end] {};
    \draw[->](theta) to node {} (A);
	\draw[->](A) to node {} (Y);
	\draw[->](X) to [out=45,in=135] node {} (Y);
\end{tikzpicture}
}
\end{center}
    \caption{Causal Graph in a randomized data collection experiment. $A$ denotes action (or treatment), $X$ context, $Y$ is the loss (or outcome). The  causal influence of the contexts on actions is done through the model $\theta$.}
    \label{fig:scm-crm}
\end{figure}
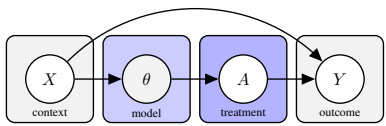

In these explanations, we recall that the distributions of contexts as well as the distribution of losses are fixed. In other words, the latter do not vary from one batch to another. In the counterfactual risk minimization framework (CRM) \citep{swaminathan2012}, the causal graph (using the conventions in \citep{PetersJanzingSchoelkopf17}) can be represented as shown in Figure \ref{fig:scm-crm}. 

\begin{figure}[htbp]
\begin{center}
\resizebox{0.3\columnwidth}{!}{
\begin{tikzpicture}[auto, thick, node distance=1cm, >=triangle 45]
\tikzstyle{unobserved}=[thick, dashed, fill=gray!0]
\tikzstyle{squared}=[thick, fill=gray!10, rounded corners=5pt]
\tikzstyle{norn}=[thick, fill=gray!0, minimum size=1cm]
\draw[squared] (-2.9,-2.7) rectangle (-1.1,0.7);
\draw[squared, fill=blue!20] (-0.9,-2.7) rectangle (0.9,0.7);
\draw[squared, fill=blue!30] (1.1,-2.7) rectangle (2.9,0.7);
\draw[squared] (3.1,-2.7) rectangle (4.9,0.7);
\draw
	node[norn] at (2,0)[causalvar](A){$A_t$}
	node[norn] at (4,0)[causalvar](Y){$Y_t$}
        node[norn] at (4,-1.5)[causalvar](Yp){$Y_{t+1}$}
	node[norn] at (-2,0)[causalvar](X){$X_t$}
        node[norn] at (-2,-1.5)[causalvar](Xp){$X_{t+1}$}
    node[squared, minimum size=1cm] at (0,0)[causalvar](theta_t){$\theta_t$}
    node[squared] at (0,-1.5)[causalvar](theta_t_p){$\theta_{t+1}$}
    node[norn] at (2,-1.5)[causalvar](A2){$A_{t+1}$};
	
\draw (0,-2.5)node[scale=0.7]{model};
\draw (2,-2.5)node[scale=0.7]{treatment};
\draw (-2.,-2.5)node[scale=0.7]{context};
\draw (4,-2.5)node[scale=0.7]{outcome};
	\draw[->](X) --  (theta_t) node[midway, near end] {};
    \draw[->](theta_t) to node {} (A);
	\draw[->](A) to node {} (Y);
	\draw[->](X) to [out=45,in=135] node {} (Y);
        \draw[->](Xp) to [out=-45,in=-135] node {} (Yp);
    \draw[->](Xp) --  (theta_t_p) node[midway, near end] {};
    \draw[->](theta_t_p) to node {} (A2);
    \draw[->](A2) to node {} (Yp);
    \draw[->](A) to node {} (theta_t_p);
    \draw[->](Y) to node {} (theta_t_p);
\end{tikzpicture}
}

\end{center}
    \caption{Causal Graph in a sequential randomized data collection experiment. $A$ denotes action (or treatment), $X$ context, $Y$ is the loss (or outcome). The contextual treatments are taken through the models $\theta_t$.}
    \label{fig:scm-scrm}
\end{figure}
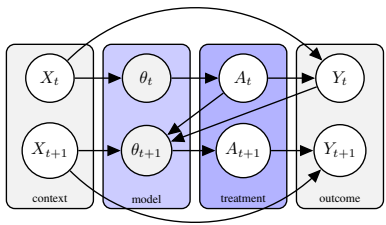

In the sequential counterfactual risk minimization (SCRM) framework, if we unfold the causal graph, the following representation can be given in Figure \ref{fig:scm-scrm}.

\begin{figure}
    \centering
    \includegraphics[width=0.6\columnwidth]{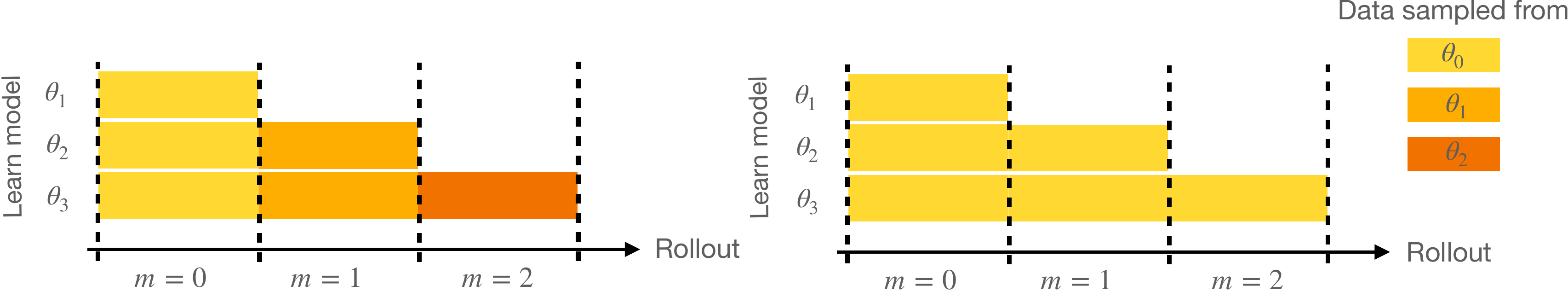}
    \caption{Graphical illustration of SCRM setup (left) and CRM (right), learned with same amount of data after each batch $m$. The training data are displayed with color block and the policy used to sample actions in these block are either adaptive (SCRM) or using the loggind model $\theta_0$ (CRM).}    \label{fig:scrm_crm_figure}
\end{figure}

Therefore, it is clear that in general, $\theta_t \notindep \theta_{t+1}$. However, from d-separation and faithfullness \citep{PetersJanzingSchoelkopf17}, we have for $t' < t$:
\[ 
    \theta_t \indep \theta_{t'} \vert \theta_{t-1}. 
\]
Therefore, given that all the dependencies are observed and that we can condition on the direct parents of a given model $\theta_t$, sequential randomized data collection are possible. We eventually provide in Figure \ref{fig:scrm_crm_figure} an illustration of SCRM and CRM.

\subsection{Multiple Importance Sampling Estimators}
\label{appendix:multi_ips}

Note that in order to avoid introducing dependencies in the excess risk bounds we analyzed, the theoretical algorithm we have studied uses geometric sample sizes to discard previous samples. However, using all past samples is effective in practice and developing guarantees for this case would be an interesting area for future research. We present in this section a estimators using aggregation of all previous information. In particular, we can use Multiple Importance Sampling (MIS) \citep{mcbook} over all previous samples. Consider in particular a partition of unity with $m > 1$ weight functions $\omega_t (a) > 0$ which satisfies $\sum_{t=0}^m \omega_{t,m} (a) = 1$ for all $a$ and $m \in \{0, \dots M \}$. The MIS estimator writes:

\begin{equation}
   \hat{L}^{\text{MIS}}_m(\theta) = \sum_{t=0}^{m}\dfrac{1}{n_t} \sum_{i=1}^{n_t} \omega_{t,m}(a_{t,i}) y_{t,i} w_{t,i}^\theta, \quad w_{t,i}^\theta = \dfrac{\pi_\theta(a_{t,i} |x_{t,i})}{\pi_{t,i}}.
       \label{eq:multi_ips}
\end{equation}

In multiple importance sampling we usually assume that the behavior distributions are independent. In our case, when we optimize $\theta_t$ based on the models $\theta_{t-1}, \dots, \theta_0$, we break this assumption. However, as we will see, we can still have the unbiasedness property and derive an estimator for the variance of the estimator.

\begin{proposition}[Unbiasedness]
The MIS estimator \eqref{eq:multi_ips} is unbiased when the loss $y$ is fixed (its distribution $\cP_{\cY}(\cdot | x, a)$ does not depend on time rollout $m$).
\end{proposition}

\begin{proof}
Let $m \in \lbrace 1, \dots M \rbrace$. We recall that at all rounds $t <m$, models $\theta_t \in \Theta$ were deployed and sets $s_t$ of $n_t$ observations $s_t~=~(x_{t, i}, a_{t, i}, l_{t,i}, \pi_{t,i})_{i = 1, \ldots, n_t}$ were collected thereof, with propensities $\pi_{t,i}~=~\pi_{\theta_t}(a_{t,i} |x_{t,i})$ to learn the next model $\theta_{t+1}$. To prove the unbiasedness we use the tower rule on the expectation and condition on previous observations $s_1, \dots s_{t-1}$:
\begin{align*}
   \displaystyle \E [\hat{L}^{\text{MIS}}_m (\theta)] &= \sum_{t=0}^{m} \dfrac{1}{n_t} \sum_{i=1}^{n_t}
   \displaystyle \E_{x, \theta_m, y} \left[ \omega_{t}(a) y w^\theta_t \right] \\
   &= \sum_{t=0}^{m}
   \displaystyle \E_{x, \theta_m, y} \left[ \omega_{t}(a) y w^\theta_t \right] \\
   &= \sum_{t=0}^{m}
   \displaystyle \E_{s_1 \dots s_{t-1}} \left[ \E_{x, \theta_m, y} \left[ \omega_{t}(a) y w^\theta_t \ | \ s_1 \dots s_{t-1} \right] \right] \\
   &= \sum_{t=0}^{m}
   \displaystyle \E_{s_1 \dots s_{t-1}} \left[ \E_{x, \theta, y} \left[ \omega_{t}(a) y  \ | \ s_1 \dots s_{t-1} \right] \right] \\
   &= \sum_{t=0}^{m}
   \displaystyle \E_{x, \theta, y} \left[ \omega_{t}(a) y \right] \\
   &= 
   \displaystyle \E_{x, \theta, y} \left[ \left(\sum_{t=0}^{m} \omega_{t}(a)  \right) y \right] \\
   &= \displaystyle \E_{x, \theta, y} \left[ y \right] \\
   &= L(\theta),
\end{align*}
where the second last line is true only when the distribution of $y$ does not change over time roll-outs $m$.
\end{proof}
Among the proposals for functions $\omega_{t} (a)$, the most 'naive' and natural heuristic is to choose 
\begin{equation}
   \omega_{t}(a) = \dfrac{n_t}{\sum_{l=1}^m n_l},
       \label{eq:naive-heuristic}
\end{equation}
which gives the naive concatenation of all IPS estimators
\begin{equation}
    \hat{L}^{\text{n-MIS}}_m(\theta) = \frac{1}{n}\sum_{t=0}^m\sum_{i=1}^{n_t} y_{t,i} \dfrac{\pi_\theta(a_{t,i}|x_{t,i})}{\pi_{\theta_t}(a_{t,i}|x_{t,i})},
\label{eq:concatenate_ips}
\end{equation}
where $n = \sum_{t=0}^m n_t$.

With the previous definition of the empirical mean estimator, we can now derive an empirical variance estimator, starting with the naive multi importance sampling estimator. We write the random variable $r^{m} = ({\pi_{\theta}}/{\pi_{\theta_m}})y$. We note that for inside a batch $m$ each realization of $r^m_i = ({\pi_{\theta}(a_{m,i} | x_{m,i})}/{\pi_{m,i}})y_{m,i}$ and $r^m_j$ are independent. But the realizations of the random variables $r^m$ and $r^{m'}$ are dependent. Writing $n=\sum_{t=0}^m n_t$
\begin{align*}
    \Var \left[ \frac{1}{n} \sum_{t=0}^m \sum_{i=1}^{n_m} r^m_i \right] &= \sum_{t=0}^m \Var \left[ \frac{1}{n} \sum_{i=1}^{n_m} r^m_i \right] + 2 \sum_{ 1\leq p<q \leq m} \Cov \left[ \frac{1}{n} \sum_{i=1}^{n_p} r^p_i, \frac{1}{n} \sum_{j=1}^{n_q} r^q_j \right] \\
    &= \frac{1}{n^2} \sum_{t=0}^m \Var \left[ \sum_{i=1}^{n_m} r^m \right] + 2 \frac{1}{n^2} \sum_{ 1\leq p<q \leq m} \sum_{i=1}^{n_p}\sum_{j=1}^{n_q} \Cov \left[   r^p,  r^q \right] \\
    &= \frac{1}{n^2} \left[ \sum_{t=0}^m \Var \left[ \sum_{i=1}^{n_m} r^m \right] + 2 \sum_{ 1\leq p<q \leq m} n_p n_q \Cov \left[   r^p,  r^q \right] \right],
\end{align*}
where the second last equality is obtained with the bilinearity of the covariance. Given the latter expression of the variance, we propose the following estimator and with a linear sampling where all $n_p=n_q$ for $p,q \in \{ 1, \dots, M \}$:
\begin{equation}
    \hat{V}^{\text{n-MIPS}}_m(\theta) = \frac{1}{n^2} \left[ \sum_{t=0}^m \hat{V}(r^t) + 2 \sum_{1 \leq p < q \leq m} n_p n_q \left( \frac{1}{n_p} \sum_{k=1}^{n_p} \big( r^p_k - \bar r_p  \big) \big( r^q_k - \bar r_q  \big)
 \right) \right],
\end{equation}
where $\hat{V}(r^m) = \frac{1}{n_m(n_m-1)} \sum_{i=1}^{n_m} \big( r_i^m - \bar r^m \big)^2$ and $\bar r^m = \frac{1}{n_m} \sum_{j=1}^{n_m} r_j^m$.

Note also that for other functions $\omega_{t} (a)$, the most studied one is the balance heuristic with $\omega_{t} \propto n_t \pi_{\theta_t}(a)$, that is:

\begin{equation}
   \omega_{t}^{BH} (a) = \dfrac{n_t \pi_{\theta_t}(a)}{\sum_{l=1}^m n_l \pi_{\theta_l}(a)}.
       \label{eq:balance-heuristic}
\end{equation}

The latter heuristic has been studied for its low variance \citep{mcbook} but these properties have been studied under an i.i.d assumption that is broken in our adaptive data collection strategy. Eventually, note that controlling the variance of this estimator with an implicit exploration estimator as we do in the i.i.d case would make a an interesting research direction.


\section{Analysis details}

\label{appendix:analysis-details}
In this section, we provide the details of our analysis by starting with essential definitions, then our proofs of variance dependent excess risk bounds and finally our regret analysis.

\subsection{Definitions}

\label{appendix:analysis_definition}

\label{sec:app-analysis-definitions}
$C_{m}(\Theta)$ is a complexity measure that will be upper-bounded by the metric entropy in sup-norm at level $\epsilon=1/n_m$ of the following function set,
\begin{equation}
    \cF_{m, \Theta} := \left\{f_\theta: (x,a,y) \in \cX \times \cA \times \cY \mapsto  \frac{1}{W}  + \frac{1}{W} y  \left(  \frac{\pi_{\theta}(a|x)}{\pi_{\theta_m}(a|x) + \alpha \pi_{\theta}(a|x)}-1 \right) \quad \text{for } \theta \in \Theta \right\} \,.
    \label{eq:function_set_pi}
\end{equation}
The latter corresponds to clipped prediction errors of policies $\pi_\theta$ normalized into $[0,1]$. More precisely, to define rigorously $C_m(\Theta)$, 
we denote for any $n_m\geq 1$ and $\epsilon >0$, the complexity of a class $\cF$ by 
\begin{equation}
  \mathcal{H}_{\infty}(\epsilon, \mathcal{F}, n) = \sup_{(x_i, a_i, y_i) \in (\cX \times \cA \times \mathcal{Y})^n} \mathcal{H}(\epsilon, \mathcal{F} \big( \{ x_i, a_i, y_i \} \big), \Vert \cdot \Vert_{\infty}) \,,
\end{equation}
where 
$
    \mathcal{F} \big( \{ x_i, a_i, y_i \} \big) = \big\{ \big(f(x_1, a_1, y_1), \dots, f(x_n, a_n, y_n)\big), f \in \mathcal{F} \big\} \subseteq \R^n
$ and the number $\mathcal{H}(\epsilon, A, \Vert \cdot \Vert_{\infty})$ is the smallest cardinality $\vert A_0\vert $ of a set $A_0 \subseteq A$ such that $A$ is contained in the finite union of $\epsilon$-balls centered at points in $A_0$ in the metric induced by $\Vert \cdot \Vert_{\infty})$.
Then, $C_m(\Theta)$ is defined by
\begin{equation}
    C_{m}(\Theta) = \log \cH_\infty(1/n_m, \cF_{m, \Theta}, 2n_m) \,.
    \label{eq:complexity}
\end{equation}

\subsection{Variance-dependent excess risk bounds}

\label{appendix:analysis-upper-bound}

We will denote by $\E_m[ \cdot] = \E[\cdot | s_0, \dots s_m]$ the conditional expectation given the set of observation samples $s_m = (x_{m,i}, a_{m,i}, y_{m,i}, \pi_{m,i})_{i=1,\dots,n_m}$ up to the rollout $m$. Here, we recall that $x_{m,i} \sim \cP_\cX$, $a_{m,i} \sim \pi_{\theta_m}(\cdot | x_{m,i})$, $y_{m,i} \sim \cP_\cY(\cdot|x_{m,i},a_{m,i})$, and $\pi_{m,i} = \pi_{\theta_m}(a_{m,i}|x_{m,i})$.  Furthermore, throughout the document, $\E_{x, \theta_m, y} \big[\cdot\big]$ (resp. $\Var_{x,\theta_m,y}\big[\cdot]$) denotes the expectation (resp. variance) in $(x,a,y)$ where $x \sim \cP_\cX$, $a \sim \pi_{\theta_m}(\cdot|x)$, and $y \sim \cP_\cY(\cdot|x,a)$.



\begin{customprop}{\ref{prop:generalization_bound}}[Generalization Error Bound]
Let $\hat{L}^{\text{IPS-IX}}_m$ and $\hat{V}^{\text{IPS-IX}}_m$ be the empirical estimators defined respectively in Eq. \eqref{eq:ips-ix} and Eq. \eqref{eq:var_ips-ix}.  Let $\delta \in (0,1)$, $\theta \in \Theta$, and $n_m \geq 2$ the number of samples associated to the logged dataset at round $m$. Then, with probability at least $1 - \delta$,
\begin{equation}
      L(\theta) \leq \hat{L}^{\text{IPS-IX}}_m(\theta) + \lambda \sqrt{\frac{\hat{V}^{\text{IPS-IX}}_m(\theta)}{n_m}} + \frac{2\lambda^2W}{n_m} + \sqrt{\frac{\log(2/\delta)}{2n_m}} \,,
\end{equation}
where $\lambda = \sqrt{18 (C_m(\Theta) + \log(2/\delta))}$. 
\end{customprop}

\begin{proof}
Let $\delta \in (0,1)$ and $\theta \in \Theta$. Since all functions in $\cF_{m, \Theta}$ defined in Eq. \eqref{eq:function_set_pi} take values in $[0,1]$, we can apply the concentration bound of~\citet[Theorem 6]{maurer2009empirical} to the set $\cF_{m, \Theta}$. This yields, with probability at least $1-\delta/2$,
\begin{equation}
    \label{eq:maurer}
    \E_{x, \theta_m, y}[f_{\theta}(x, a, y)] - \frac{1}{n_m} \sum_{i=1}^{n_m} f_{\theta}(x_{m,i}, a_{m,i}, y_{m,i}) \leq \sqrt{\frac{18\hat{V}_{n_m}(f_\theta)(C_m(\Theta) + \log(2/\delta))}{n_m}} + \frac{15 (C_m(\Theta) + \log(1/\delta))}{(n_m - 1)} \,,
\end{equation}
where 
$$
\hat{V}_{n_m}(f_\theta) = \frac{1}{n_m-1} \sum_{i=1}^{n_m} \Big( f_{\theta}(x_{m, i}, a_{m, i}, y_{m,i}) - \frac{1}{n_m} \sum_{j=1}^{n_m} f_{\theta}(x_{m, j}, a_{m, j}, y_{m,j}) \Big)^2
$$
is an estimation of the sample variance. Let $\alpha >0$ and define the following biased estimator of the excess risk: 
\begin{equation}
  L^\alpha_m(\theta) = \mathbb E_{x, \theta_m, y} \left[ y \left(  \frac{\pi_{\theta}(a|x)}{\pi_{\theta_m}(a|x) + \alpha \pi_{\theta}(a|x)}-1 \right) \right] \qquad \forall \theta \in \Theta.
  \label{eq:clipped_expected_risk}
\end{equation}
We recall that $\E_{x, \theta_m, y} \big[\cdot\big]$ denotes the expectation in $(x,a,y)$ where $x \sim \cP_\cX$, $a \sim \pi_{\theta_m}(\cdot|x)$, and $y \sim \cP_\cY(\cdot|x,a)$. 
By construction of $f_\theta$ (see Eq.~\eqref{eq:function_set_pi}),
\begin{align*}
    \E_{x, \theta_m, y}[f_{\theta}(x, a, y)]    &  = \frac{1}{W} + \frac{1}{W} L^\alpha_m(\theta) \\
    \frac{1}{n_m} \sum_{i=1}^{n_m} f_{\theta}(x_{m, i}, a_{m,i}, y_{m,i}) 
        & = \frac{1}{W} + \frac{1}{W} \hat{L}^{\text{IPS-IX}}_m(\theta) - \frac{1}{W n_m} \sum_{i=1}^{n_m} y_{m,i} \\
    \hat{V}_{n_m}(f_\theta) &  = \frac{1}{W^2} \hat{V}^{\text{IPS-IX}}_m(\theta) \,,
\end{align*}
where $\hat{L}^{\text{IPS-IX}}_m$ and $\hat{V}^{\text{IPS-IX}}_m$ are defined respectively in Eq. \eqref{eq:ips-ix} and Eq. \eqref{eq:var_ips-ix}.
Thus, multiplying~\eqref{eq:maurer} by~$W$, substituting the above terms, and using $\lambda = \sqrt{18 (C_m(\Theta) + \log(2/\delta))}$, yields 
\begin{equation}
    L^\alpha_m(\theta) - \hat{L}^{\text{IPS-IX}}_m(\theta) + \frac{1}{n_m} \sum_{i=1}^{n_m} y_{m,i} \leq \lambda \sqrt{\frac{\hat{V}^{\text{IPS-IX}}_m(\theta)}{n_m}} + \frac{15\lambda^2W}{18(n_m-1)},
    \label{eq:CI_clipped_expectation}
\end{equation} 
with probability~$1 - \delta/2$.
Now, let us decompose
\begin{equation*}
    L^\alpha_m(\theta) = \mathbb E_{x, \theta_m, y} \left[ y \left(  \frac{\pi_{\theta}(a|x)}{\pi_{\theta_m}(a|x) + \alpha \pi_{\theta}(a|x)}-1 \right) \right] = \mathbb E_{x, \theta_m, y} \left[ y   \frac{\pi_{\theta}(a|x)}{\pi_{\theta_m}(a|x) + \alpha \pi_{\theta}(a|x)} \right] - L(\theta_m).
\end{equation*}
But, since the losses $y$ are bounded in $[-1, 0]$ almost surely, 
\begin{equation*}
    \mathbb E_{x, \theta_m, y} \left[ y   \frac{\pi_{\theta}(a|x)}{\pi_{\theta_m}(a|x) + \alpha \pi_{\theta}(a|x)} \right] \geq \mathbb E_{x, \theta_m, y} \left[ y   \frac{\pi_{\theta}(a|x)}{\pi_{\theta_m}(a|x)} \right] = L(\theta),
\end{equation*}
which, substituted into the previous equation, entails,
\begin{equation}
     L^\alpha_m(\theta) \geq L(\theta) - L(\theta_m).
\label{eq:lower_bound_clipped_expected_risk}
\end{equation}
Lower-bounding the left-hand side of~\eqref{eq:CI_clipped_expectation}, we thus get w.p $1-\delta/2$,
\[
    L(\theta) - \hat{L}^{\text{IPS-IX}}_m(\theta) \leq \lambda \sqrt{\frac{\hat{V}^{\text{IPS-IX}}_m(\theta)}{n_m}} + \frac{15\lambda^2W}{18(n_m-1)} + L(\theta_m) -  \frac{1}{n_m} \sum_{i=1}^{n_m} y_{m,i}.
\]
Using $\E_{m-1}[y_{m,i}] = L(\theta_m)$ and applying Hoeffding's inequality, this further yields w.p. $1-\delta$
\begin{equation}
    L(\theta) \leq \hat{L}^{\text{IPS-IX}}_m(\theta) + \lambda \sqrt{\frac{\hat{V}^{\text{IPS-IX}}_m(\theta)}{n_m}} + \frac{15\lambda^2W}{18(n_m-1)} + \sqrt{\frac{\log(2/\delta)}{2n_m}}.
    \label{eq:CI_clipped_expectation}
\end{equation}
Eventually, note that $(n_m-1)^{-1} \leq ({2}/{n_m})$ since $n_m \geq 2$. Thus, 
\begin{equation}
    L(\theta) \leq \hat{L}^{\text{IPS-IX}}_m(\theta) + \lambda \sqrt{\frac{\hat{V}^{\text{IPS-IX}}_m(\theta)}{n_m}} + \frac{2\lambda^2W}{n_m} + \sqrt{\frac{\log(2/\delta)}{2n_m}},
    \label{eq:CI_clipped_expectation}
\end{equation}

which concludes the proof.





\end{proof}

\begin{customprop}{\ref{prop:excess_risk}}[Conservative Excess Risk]
Let $m\geq 0$ and $\theta_m \in \Theta$. Let $s_m = (x_{m,i},a_{m,i},y_{m,i},\pi_{m,i})_{1\leq i\leq n_m}$ be a set of samples collected with $a_{m,i} \sim \pi_{\theta_m}(\cdot|x_{m,i})$. Then, under Assumptions~\ref{assum:bounded_ipw} and~\ref{assumption:logarithmic_complexity}, the solution $\theta_{m+1}$ of Problem~\eqref{eq:learning_objective_scrm} with the IPS-IX estimator in Eq. \eqref{eq:conservative_learning} on the samples $s_m$ satisfies the excess risk upper-bound
\begin{equation}
    \Delta_{m+1} =~L(\theta_{m+1}) - L(\theta^*) \lesssim  \sqrt{\frac{d\log(n_m) + \log(1/\delta)}{n_m} \nu_m^2}  + \frac{W^2 + W (d \log(n_m) + \log(1/\delta))}{n_m},
\end{equation}
where $\nu_m^2 = \Var_{x, \theta_{m}}\left(\dfrac{\pi_{\theta^*}(a|x)}{\pi_{\theta_{m}}(a|x)}\right)$.
\end{customprop}

\begin{proof}
We consider the notations of the proof of Proposition~\ref{prop:generalization_bound}. Fix $\theta^* \in \Theta$. Applying,  Theorem 15 of \cite{maurer2009empirical}\footnote{Note that in their notation, $\log \mathcal{M}_n(\pi)$ equals $ C_m(\Theta) + \log(10)$, $\bf X$ is the dataset $\smash{\{(x_i,a_i,y_i)\}_{1\leq i\leq n}}$ where $\smash{(x_i,a_i,y_i) \stackrel{i.i.d.}{\sim} \cP_{\cX}\times \pi_{\theta_m}(\cdot|x) \times \cP_{\cY}(\cdot | a, x)}$, and $P(\cdot, \mu)$ is the expectation with respect to one test sample $\E_{x, \theta_m, y} [\,\cdot\,]$.} to the function set $\cF_{m, \Theta}$ defined in~\eqref{eq:function_set_pi}, we get with probability $1-\delta$
\begin{multline*}
    \E_{x, \theta_m, y}[f_{\theta_{m+1}}(x, a, y)] - \E_{x, \theta_m, y}[f_{\theta^*}(x, a, y)]\\
    \leq \sqrt{\frac{32 \Var_{x, \theta_m, y} \big[f_{\theta^*}(x,a, y)\big] \big( C_m(\Theta) + \log \frac{30}{\delta}\big)}{n_m} } + \frac{22\big( C_m(\Theta) + \log\frac{30}{\delta}\big)}{n_m-1} \,.
\end{multline*}
This can be written as:
\begin{equation}
    \Delta^*_{m} \leq U^*_m,
\label{eq:delta_u_m}
\end{equation}
with the following definitions:
\begin{align}
    \Delta^*_{m} &= \E_{x, \theta_m, y}[f_{\theta_{m+1}}(x, a, y)] - \E_{x, \theta_m, y}[f_{\theta^*}(x, a, y)] \nonumber \\ U^*_m &= 
     \sqrt{\frac{32 \Var_{x, \theta_m, y} \big[f_{\theta^*}(x,a,y)\big] \big( C_m(\Theta) + \log \frac{30}{\delta}\big)}{n_m} } + \frac{22\big( C_m(\Theta) + \log\frac{30}{\delta}\big)}{n_m-1}. \label{eq:Um}
\end{align}


\paragraph{Step: Lower bounding $\Delta^*_{m}$}

Using the definition of $f_{\theta}(x,a, y)$ in~\eqref{eq:function_set_pi} and that of $L^\alpha_m$ in Eq.~\eqref{eq:clipped_expected_risk}, we have
\begin{equation*}
    \E_{x, \theta_m, y}[f_{\theta_{m+1}}(x, a, y)] = \frac{1}{W} + \frac{1}{W} L^\alpha_m(\theta_{m+1}).
\end{equation*}
Thus, $\Delta_m^*$ can be re-written as
\begin{equation*}
    \Delta^*_{m} = \frac{1}{W} \left( L^\alpha_m(\theta_{m+1}) - L^\alpha_m(\theta^*) \right)\,,
\end{equation*}
which we now lower-bound. To do so, we begin by upper-bounding $L^{\alpha}_m(\theta^*)$. It can be expressed as
\begin{equation}
 L^\alpha_m(\theta^*) = \mathbb E_{x, \theta_m, y} \left[ y   \frac{\pi_{\theta^*}(a|x)}{\pi_{\theta_m}(a|x) + \alpha \pi_{\theta^*}(a|x)} \right]  - L(\theta_m).
 \label{eq:biasbound}
\end{equation}
To shorten notation, from now on and throughout this proof, we write $\pi_\theta$ instead of $\pi_\theta(a|x)$, omitting the dependence on $a$ and $x$. Using the inequality $(1+x)^{-1} \geq 1-x$ for $x\geq 0$, we have
\begin{align}
    \mathbb E_{x, \theta_m, y} \left[ y   \frac{\pi_{\theta^*}}{\pi_{\theta_m} + \alpha \pi_{\theta^*}} \right] 
    & = \mathbb E_{x, \theta_m, y} \left[ y  \frac{\pi_{\theta^*}}{\pi_{\theta_m}} \frac{1}{1 + \alpha \frac{\pi_{\theta^*}}{\pi_{\theta_m}}} \right] \label{eq:factorize_importance_weight} \\
    &\leq  \mathbb E_{x, \theta_m, y} \left[ y  \frac{\pi_{\theta^*}}{\pi_{\theta_m}} \right] - \alpha \mathbb E_{x, \theta_m, y} \left[ y \left( \frac{\pi_{\theta^*}}{\pi_{\theta_m}} \right)^2 \right] \nonumber \nonumber \\
    &=  L(\theta^*) - \alpha \mathbb E_{x, \theta_m, y} \left[ y \left( \frac{\pi_{\theta^*}}{\pi_{\theta_m}} \right)^2 \right] \nonumber \\
    & \leq L(\theta^*) + \alpha W^2 \,,
\label{eq:limited_developement}
\end{align}
where the last inequality is by Assumption~\ref{assum:bounded_ipw} and because $y \in [-1, 0]$.
Together with~\eqref{eq:biasbound}, we get
\begin{equation*}
    L^\alpha_m(\theta^*)  \leq L(\theta^*) + \alpha W^2 - L(\theta_m).
\end{equation*}
We recall that $L(\theta_{m+1}) - L(\theta_m) \leq L^\alpha_m(\theta_{m+1})$ by Eq.\eqref{eq:lower_bound_clipped_expected_risk}. Therefore, 
\begin{equation*}
    \frac{1}{W}(L(\theta_{m+1}) - L(\theta^*) -\alpha W^2)  \leq \frac{1}{W} \left( L^\alpha_m(\theta_{m+1}) - L^\alpha_m(\theta^*) \right),
\end{equation*}
which finally gives
\begin{equation}
    \frac{1}{W}(L(\theta_{m+1}) - L(\theta^*) -\alpha W^2)  \leq \Delta^*_{m}.
\label{eq:lowerbound_deltam}
\end{equation}

\paragraph{Step: Upper bound $U^*_m$} 
By definition of $f_{\theta}(x,a, y)$ in~\eqref{eq:function_set_pi}, we have
\begin{multline*}
    \Var_{x, \theta_m, y} \big[f_{\theta^*}(x, a, y)\big]  = \frac{1}{W^2} \Var_{x, \theta_m, y} \left[ y \left(  \frac{\pi_{\theta^*}}{\pi_{\theta_m} + \alpha \pi_{\theta^*}}-1 \right) \right] \\
     \leq \frac{1}{W^2} \mathbb E_{x, \theta_m, y} \left[ y^2 \left(  \frac{\pi_{\theta^*}}{\pi_{\theta_m} + \alpha \pi_{\theta^*}}-1 \right)^2 \right] 
     \leq \frac{1}{W^2} \mathbb E_{x, \theta_m} \left[ \left(  \frac{\pi_{\theta^*}}{\pi_{\theta_m} + \alpha \pi_{\theta^*}}-1 \right)^2 \right].
\end{multline*}
Then, using the inequality $(x+y)^2 \leq 2x^2+2y^2$, for $x,y \in \R$, this may be upper-bounded as
\begin{multline}
 \Var_{x, \theta_m, y} \big[f_{\theta^*}(x, a, y)\big]  \\
 \leq \frac{2}{W^2} \mathbb E_{x, \theta_m} \left[ \left(  \frac{\pi_{\theta^*}}{\pi_{\theta_m} + \alpha \pi_{\theta^*}}-\mathbb E_{x, \theta_m} \left[   \frac{\pi_{\theta^*}}{\pi_{\theta_m} + \alpha \pi_{\theta^*}}\right] \right)^2 \right] + \frac{2}{W^2} \left( \E_{x, \theta_m} \left[   \frac{\pi_{\theta^*}}{\pi_{\theta_m} + \alpha \pi_{\theta^*}}\right] -1 \right)^2.
  \label{eq:twoterms}
\end{multline}
On the one hand, the first term of the right-hand side may be upper-bounded as
\begin{equation*}
    \mathbb E_{x, \theta_m} \left[ \left(  \frac{\pi_{\theta^*}}{\pi_{\theta_m} + \alpha \pi_{\theta^*}}-\mathbb E_{x, \theta_m} \left[   \frac{\pi_{\theta^*}}{\pi_{\theta_m} + \alpha \pi_{\theta^*}}\right] \right)^2 \right] = \Var_{x, \theta_m} \left[ \frac{\pi_{\theta^*}}{\pi_{\theta_m} + \alpha \pi_{\theta^*}} \right] \leq \nu_m^2,
\end{equation*}
where $\nu_m^2 = \Var_{x, \theta_m} \left[ \frac{\pi_{\theta^*}}{\pi_{\theta_m}} \right]$. On the other hand, for the second term, we use the same factorization as in Eq.~\eqref{eq:factorize_importance_weight} to get 
\[-\alpha 
 \E_{x, \theta_m} \left[   \left( \frac{\pi_{\theta^*}}{\pi_{\theta_m} }\right)^2\right] \leq \E_{x, \theta_m} \left[   \frac{\pi_{\theta^*}}{\pi_{\theta_m} + \alpha \pi_{\theta^*}}\right] -1 \leq 0 \,,
 \]
which yields the upper-bound
\begin{align*}
    \left( \E_{x, \theta_m} \left[   \frac{\pi_{\theta^*}}{\pi_{\theta_m} + \alpha \pi_{\theta^*}}\right] -1 \right)^2 &\leq \alpha^2  \E_{x, \theta_m} \left[   \left( \frac{\pi_{\theta^*}}{\pi_{\theta_m} }\right)^2\right] 
    \leq \alpha^2  W^2 .
\end{align*}
Therefore, substituting the last two upper-bounds into~\eqref{eq:twoterms} entails
\begin{equation*}
    \Var_{x, \theta_m, y} \big[f_{\theta^*}(x, a, y)\big]  \leq \frac{2}{W^2} \big(\nu_m^2 + \alpha^2 W^2  \big) \,.
\end{equation*}
Then, replacing this upper-bound into the definition of $U_m^*$ in~\eqref{eq:Um} and using Assumption \ref{assumption:logarithmic_complexity} to upper bound the terms in $C_m(\Theta) \leq d \log(n_m)$, we obtain the following upper-bound
\begin{align}
    U^*_m &
        \leq \frac{1}{W} \sqrt{\frac{64 (\nu_m^2 + \alpha^2 W)  \big( d \log(n_m)  + \log \frac{30}{\delta}\big)}{n_m} } + \frac{22\big( d \log(n_m) + \log\frac{30}{\delta}\big)}{n_m-1} \nonumber \\
        & \leq \frac{1}{W} \sqrt{\frac{64 (\nu_m^2 + \alpha^2 W)  \big( d \log(n_m)  + \log \frac{30}{\delta}\big)}{n_m} } + \frac{44\big( d \log(n_m) + \log\frac{30}{\delta}\big)}{n_m}\,,
    \label{eq:upperbound_um}
\end{align}
where the last inequality is because $n_m \geq 2$. 

\paragraph{Step: excess risk upper bound}
Setting $\alpha = \frac{1}{n_m}$ and using the two previous bounds \eqref{eq:lowerbound_deltam} and \eqref{eq:upperbound_um} respectively on $\Delta^*_m$ and on $U^*_m$ into~\eqref{eq:delta_u_m}, we get
\begin{equation}
    L(\theta_{m+1}) - L(\theta^*) \leq  \sqrt{\frac{64 \big( d \log(n_m)  + \log \frac{30}{\delta}\big)}{n_m} \big(\nu_m^2 + \frac{1}{n_m^2} W^2\big)} + W \frac{44\big( d \log(n_m) + \log\frac{30}{\delta}\big)}{n_m} + \frac{1}{n_m}  W^2.
\label{eq:excess_risk_complete_bound1}
\end{equation}
Using that $\sqrt{a + b} \leq \sqrt{a} + \sqrt{b}$, we have that
\begin{equation*}
    \sqrt{\frac{64 \big( d \log(n_m)  + \log \frac{30}{\delta}\big)}{n_m} \big(\nu_m^2 + \frac{1}{n_m^2} W^2\big)   } \leq \sqrt{\frac{64 \big( d \log(n_m)  + \log \frac{30}{\delta}\big)}{n_m} \nu_m^2}  + \frac{W}{n_m}\sqrt{\frac{64 \big( d \log(n_m)  + \log \frac{30}{\delta}\big)}{n_m}}.
\end{equation*}
Then, since $n_m \geq 2$ and $\delta <1$, we have $d\log(n_m) + \log(30/\delta) \geq \log(2) + \log(30) \geq 4$, which yields
\begin{equation*}
    \frac{1}{n_m}\sqrt{\frac{64 \big( d \log(n_m)  + \log \frac{30}{\delta}\big)}{n_m}} \leq \frac{\sqrt{32 \big( d \log(n_m)  + \log \frac{30}{\delta}\big)}}{n_m} \leq \frac{\sqrt{8}  \big( d \log(n_m)  + \log \frac{30}{\delta}\big)}{n_m}.
\end{equation*}
Substituting the last two inequalities into~\eqref{eq:excess_risk_complete_bound1} finally entails
\begin{equation}
    L(\theta_{m+1}) - L(\theta^*) \leq  8 \sqrt{\frac{d \log(n_m)  + \log \frac{30}{\delta}}{n_m} \nu_m^2} + 47 W \frac{d \log(n_m) + \log\frac{30}{\delta}}{n_m} + \frac{W^2}{n_m},
\label{eq:excess_risk_complete_bound}
\end{equation}
which concludes the proof.
\end{proof}

\subsection{Regret analysis}

\label{appendix:regret_proof}

\begin{customprop}{\ref{prop:regret_bound}}[Regret upper-bound]
    Let $n_0, n \geq 2$ and $\theta^* \in \argmin_{\theta} L(\theta)$. Let $n_m = n_02^m$ for $m=0,\dots,M  = \big\lfloor \log_2 ( 1+ \frac{n}{n_0}) \big\rfloor$. Then, under Assumptions~\ref{assum:bounded_ipw},~\ref{assumption:logarithmic_complexity} and~\ref{assum:holderian_error_bound},  the SCRM procedure (Alg. \ref{alg:scrm}) satisfies the excess risk upper-bound
   \begin{equation*}
       L(\theta_M) - L(\theta^*) \leq O\Big( n^{-\frac{1}{2-\beta}}\log n\Big) \,.
   \end{equation*}
   Moreover, the expected regret is upper-bounded as follows:
   \begin{equation*}
    R_n = \E\bigg[ \sum_{m=0}^M n_{m+1} \big(L(\theta_m) - L(\theta^*)\big) \bigg] \leq O\Big( n^{\frac{1-\beta}{2-\beta}}\log(n)^2\Big) .
\end{equation*}

\end{customprop}

\begin{proof}
First, note that for $n_m = n_02^m$ and $M  = \big\lfloor \log_2 ( 1+ \frac{n}{n_0}) \big\rfloor$, we have
$
    \sum_{m=0}^{M-1} n_m = n_0(2^M - 1) \leq n. 
$
Hence, Alg.~\ref{alg:scrm} has collected at most $n$ samples to design the estimator $\theta_M$. 
For $m\geq 0$, we recall $\Delta_m = L(\theta_{m}) - L(\theta^*)$ and use Eq.~\eqref{eq:excess_risk_complete_bound} to write
\begin{align}
     \Delta_{m+1} 
        & \leq  8 \sqrt{\frac{d \log(n_m)  + \log \frac{30}{\delta}}{n_m} \nu_m^2} + 47 W \frac{d \log(n_m) + \log\frac{30}{\delta}}{n_m} + \frac{W^2}{n_m} \nonumber \\
        & \leq  8 \sqrt{\frac{d \log(n)  + \log \frac{30}{\delta}}{n_m} \nu_m^2} + 47 W \frac{d \log(n) + \log\frac{30}{\delta}}{n_m} + \frac{W^2}{n_m} \nonumber \\
        & = C \sqrt{\frac{\nu_m^2}{n_m}} + \frac{B}{n_m} \label{eq:simplified_excess_risk} \,,
\end{align}
where $C = 8 \sqrt{d \log(n)  + \log \frac{30}{\delta}}$ and $B = W^2 + 47 W (d \log(n) + \log\frac{30}{\delta})$ are independent of $m$.

\paragraph{Step: Obtaining a recurrence relation for $\Delta_{m+1}$}

By Assumption \ref{assum:holderian_error_bound}, there exist $\gamma>0$ and $\beta \in [0,1]$ such that
\begin{equation*}
    \nu_m^2 = \Var_{x, \theta_m}\left(\frac{\pi_{\theta^*}}{\pi_{\theta_m}} \right) \leq \frac{1}{\gamma} \big(L(\theta_m) - L(\theta^*)\big)^\beta = \frac{\Delta_m^\beta}{\gamma} \,.
\end{equation*}
Replacing $\nu_m^2$ in Eq.~\eqref{eq:simplified_excess_risk} thus entails
\begin{align}
    \Delta_{m+1} 
        & \leq C \sqrt{\frac{1}{\gamma}\frac{\Delta_m^\beta}{n_m}} + \dfrac{B}{n_m} \nonumber \\
        & \leq  C  2^{-\frac{m}{2}} \sqrt{\frac{n_0}{\gamma} \Delta_{m}^\beta} + B 2^{-m} n_0  \qquad \leftarrow{n_m = n_0 2^m} \nonumber \\
        & = C \sqrt{\frac{n_0}{\gamma}} 2^{-\frac{m}{2}}   \Delta_{m}^{\beta/2} + B 2^{-m} n_0 \,.\label{eq:recurrence_delta_m} 
\end{align}

\paragraph{Step: Solving the recurrence relation for $\Delta_{m}$}
 We then insure by induction that $\Delta_m$ satisfies 
\begin{equation}
    \Delta_m \leq  c_0 2^{\frac{-m}{2-\beta}} \,,
\label{eq:delta_m}
\end{equation}
for some $c_0 >0$ that will be specified by the analysis.

\textbf{Base step} Since losses take values in $[-1,0]$, $\Delta_0 = L(\theta_0) - L(\theta^*) \leq 1$. Equation~\eqref{eq:delta_m} is thus satisfied for $m=0$ as soon as $c_0 \geq 1$.

\textbf{Induction step} Let $m\geq 0$. We assume that $\Delta_m \leq 
 c_0 2^{\frac{-m}{2-\beta}}$ and prove Equation~\eqref{eq:delta_m} for $\Delta_{m+1}$.  Using Eq. \eqref{eq:recurrence_delta_m}, we have
\begin{align}
    \Delta_{m+1} &\leq C \sqrt{\frac{n_0}{\gamma}} 2^{-\frac{m}{2}}   \Delta_{m}^{\beta/2} + B 2^{-m} n_0  \nonumber \\
        & \leq C \sqrt{\frac{n_0}{\gamma}} 2^{-\frac{m}{2}}   {c_0}^{\beta/2} 2^{-\frac{m\beta}{\beta(2-\beta)}} + B 2^{-m} n_0  \qquad \leftarrow \quad \text{by induction} \nonumber \\
        & \leq \max\Big\{2 C \sqrt{\frac{n_0}{\gamma}}   {c_0}^{\frac{\beta}{2}} 2^{-\frac{m}{2} - \frac{m}{2-\beta}}, 2B 2^{-m} n_0\Big\} \,.\label{eq:induction_deltam}
\end{align}
Now, we show that both terms inside the maximum can be upper-bounded by $c_0 2^{-(m+1)/(2-\beta)}$ as soon as $c_0$ is large enough. On the one hand, if $c_0 \geq 4B n_0$, we have
\[
    2B 2^{-m} n_0 \leq c_0 2^{-(m+1)} \leq c_0 2^{-\frac{m+1}{2-\beta}} \,.
\]
On the other hand, if $c_0 \geq (4 C^2 n_0/\gamma)^{1/(2-\beta)}$, we also have
\[
    2 C \sqrt{\frac{n_0}{\gamma}}   {c_0}^{\frac{\beta}{2}} 2^{-\frac{m}{2} - \frac{m}{2-\beta}} \leq 2 C \sqrt{\frac{n_0}{\gamma}}   {c_0}^{\frac{\beta}{2}}  2^{- \frac{m+1}{2-\beta}}  \leq c_0 2^{- \frac{m+1}{2-\beta}} \,.
\]
Combining the above two upper-bounds with~\eqref{eq:induction_deltam} concludes the induction step under the condition
\[
    c_0 \geq \max \bigg\{1,  \Big(\frac{4 C^2 n_0}{\gamma}\Big)^{\frac{1}{2-\beta}}, 4 B n_0 \bigg\} \,.
\]
\paragraph{Step: conclusion} Finally, setting the above value for $c_0$ we proved that for all $m\geq 0$, we have
\begin{align}
    \Delta_m & \leq \max \bigg\{1,  \Big(\frac{4 C^2 n_0}{\gamma}\Big)^{\frac{1}{2-\beta}}, 4 B n_0 \bigg\} 2^{- \frac{m}{2-\beta}} \nonumber \\
        & \leq  \bigg(1+  \Big(\frac{4C^2 n_0}{\gamma}\Big)^{\frac{1}{2-\beta}} +  4B n_0 \bigg) 2^{- \frac{m}{2-\beta}} \nonumber \\
        & =   \bigg(1+ \bigg(\frac{256 (d \log n  + \log \frac{30}{\delta}) n_0}{\gamma}\bigg)^{\frac{1}{2-\beta}} +  W^2n_0 + 47 W n_0 \Big(d \log n + \log\frac{30}{\delta}\Big) \bigg) 2^{- \frac{m}{2-\beta}} \label{eq:upperbound_deltam} \,,
\end{align}
where the last equality is by substituting the values of $B$ and $C$ from~\eqref{eq:simplified_excess_risk}. For the final step $M~=~\lfloor \log_2(\frac{n}{n_0}+1) \rfloor$, this yields
\begin{align*}
    \Delta_{M} 
        & \leq \bigg(1+ \bigg(\frac{256 (d \log n  + \log \frac{30}{\delta}) n_0}{\gamma}\bigg)^{\frac{1}{2-\beta}} +  W^2n_0 + 47 W n_0 \Big(d \log n + \log\frac{30}{\delta}\Big) \bigg) 2^{- \frac{M}{2-\beta}} \\
        & \leq 2 \bigg(1+ \bigg(\frac{256 (d \log n  + \log \frac{30}{\delta}) n_0}{\gamma}\bigg)^{\frac{1}{2-\beta}} +  W^2n_0 + 47 W n_0 \Big(d \log n + \log\frac{30}{\delta}\Big) \bigg) \times \Big(\frac{n_0}{n}\Big)^{\frac{1}{2-\beta}} \\
        & = O\Big( n^{-\frac{1}{2-\beta}}\log n\Big) \,. 
\end{align*}
This concludes the first part of the proof.

\paragraph{Regret upper-bound}
To upper bound the cumulative regret, using $n_{m+1} = n_0 2^{m+1}$, we write
\begin{align*}
    R_n & = \sum_{m=0}^{M} \Delta_m n_{m+1} 
         \stackrel{\eqref{eq:upperbound_deltam}}{\leq} D \sum_{m=0}^M  2^{-\frac{m}{2-\beta}} n_{m+1} = 2 D n_0 \sum_{m=0}^M  2^{\big(\frac{1-\beta}{2-\beta}\big) m } \,,
\end{align*}
where 
\[
    D = 1+ \bigg(\frac{256 (d \log n  + \log \frac{30}{\delta}) n_0}{\gamma}\bigg)^{\frac{1}{2-\beta}} +  W^2n_0 + 47 W n_0 \Big(d \log n + \log\frac{30}{\delta}\Big)  \,.
\]
Then, computing the sum for $M~=~\lfloor \log_2(\frac{n}{n_0}+1) \rfloor$, we have
\[
   R_n\leq  2 D n_0  \sum_{m=0}^M  2^{\big(\frac{1-\beta}{2-\beta}\big) m } 
    \leq  2 D n_0  (M+1) 2^{\big(\frac{1-\beta}{2-\beta}\big) M } \leq 
     2 D n_0 \Big(1 + \log_2\Big(\frac{n}{n_0}+1\Big) \Big) \times  \Big(1 + \frac{n}{n_0}\Big)^{\frac{1-\beta}{2-\beta}} \,.
\]
Using that $D = O(\log n)$, we finally obtain
\[
    R_n \leq O\Big(n^{\frac{1-\beta}{2-\beta}} \log(n)^2\Big) \,.
\]
\end{proof}

\section{Additional discussions on the Hölderian Bound Assumption~\ref{assum:holderian_error_bound}}

In this appendix, we discuss Assumption~\ref{assum:holderian_error_bound} on different particular examples. 

\subsection{Verification of the assumption on a toy example with Gaussian families}

\label{app:gaussian_example}
\begin{figure}[h]
    \centering
    \includegraphics[width=0.45\linewidth]{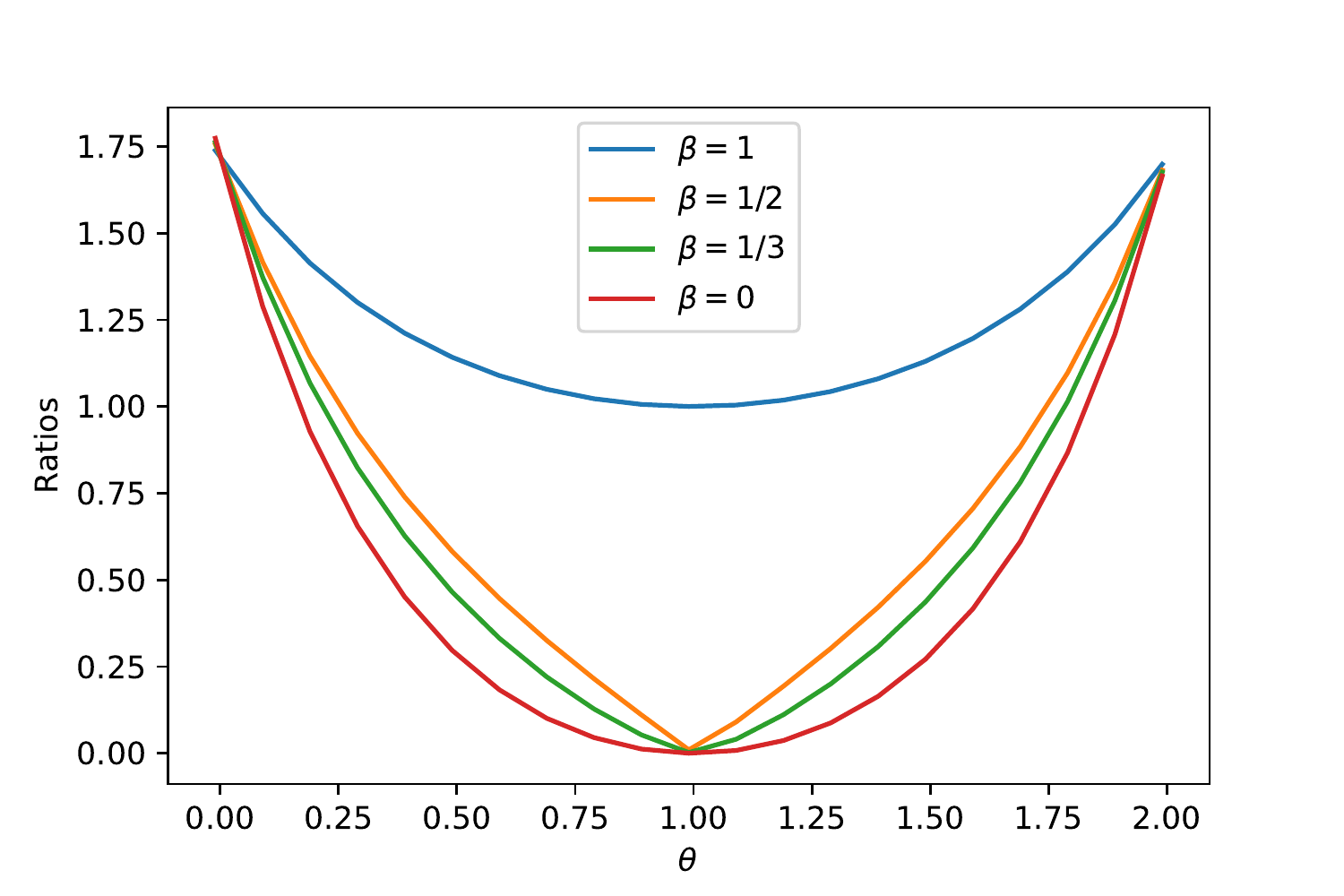}
    \caption{Ratio $R$ defined in \eqref{eq:analytical_ratio} with different values of $\beta$.}
    \label{fig:heb}
\end{figure}

We consider the setting of Example \ref{example:gaussian_policies}. In the latter,  the policies are Gaussian of the form $\pi_\theta = \mathcal{N}(\theta, \sigma^2)$ and the loss is defined by $l_t(a)=(a-y_t)^2-1$ where $y_t \sim \mathcal{N}(\theta^*, \sigma^2)$. There is no loss in generality in assuming $\sigma^2 = 1$. Then, we can compute 
\[
     L(\theta) - L(\theta^*) = ( \theta - \theta^* )^2 \qquad \text{ and }  \qquad \Var_{\theta}\left[\frac{\pi_{\theta^*}(a)}{\pi_{\theta}(a)} \right] = \exp \left( (\theta^* - \theta)^2 \right) - 1 \,.
\]
We recall that we are interested in verifying the existence of $\gamma>0$ and $\beta > 0$ for which Assumption \ref{assum:holderian_error_bound} holds, that is in this case for any $\theta \in \Theta$:
\begin{equation}
    \gamma \Var_{\theta} \left[ \dfrac{\pi_{\theta^*}(a)}{\pi_{\theta}(a)} \right] \leq  \big( L(\theta) - L(\theta^*) \big)^{\beta}\,,
    \label{eq:heb_example}
\end{equation}
which may be re-written here as
\[
     \gamma \big(\exp \left( (\theta^* - \theta)^2 \right) - 1\big) \leq (\theta - \theta^*)^{2\beta} \,.
\]
The latter is satisfied for any $\beta \leq 1$ as soon as $\Theta$ is a bounded interval. Note that the constant $\gamma$ may decrease exponentially fast as the diameter of $\Theta$ increases. To illustrate, the existence of such couples $(\beta, \gamma)$, we plot in Fig. \ref{fig:heb} different values of the following ratio
\begin{equation}
    R(\theta, \beta)  = \dfrac{\Var_{\theta} \left[ \dfrac{\pi_{\theta^*}(a)}{\pi_{\theta}(a)} \right]}{\big( L(\theta) - L(\theta^*) \big)^{\beta}} 
    = \dfrac{\exp \left( (\theta^* - \theta)^2 \right) - 1}{\big(\Vert \theta - \theta^* \Vert^2\big)^\beta}.
    \label{eq:analytical_ratio}
\end{equation}
The value of $\gamma$ can be found for different values of $\beta$ in Fig. \ref{fig:heb} by taking $\frac{1}{\gamma} = \max_{\theta} R(\theta, \beta)$.
Higher values of $\beta$ induce faster rates and lower values of $\gamma$ induce worst constant terms in the excess risk and regret bounds. Eventually, note that SCRM does not need those parameters to run and those two parameters $\gamma, \beta$ are automatically calibrated by SCRM to find the best trade-off. 

\subsection{Discussion of Assumption~\ref{assum:holderian_error_bound} for Exponential Families}

\label{appendix:exponential_families}

In this section, we consider a more realistic example in which policies belong to an exponential family.  
That is, we assume that the policies are parameterized by a parameter $\eta \in \mathbb{R}^q$ and can be written in the form: 
\begin{equation*}
  \forall a \in \cA,\qquad   \pi_\eta(a) = e^{\eta \cdot t(a) - A(\eta)} h(a),
\end{equation*}
for some known function $h:\cA \to \R_+$ and  sufficient statistic $t:\cA \to \R^q$. 
Here, $A(\eta)$ is a normalization constant, so that $e^{A(\eta)} = \int_a e^{\eta \cdot t(a)} h(a) \diff a$. 
We provide in Example~\ref{ex:expfamily} a concrete example considered by~\citep{swaminathan2012,faury2020distributionally}.
To ease the notation, we removed here the dependency on contexts, but the generalization to contextual policies can be made similarly. The importance weight ratio may be written as,
\begin{equation}
    \frac{\pi_\eta(a)}{\pi_{\eta_m}(a)}= e^{\left(\eta - \eta_m \right) t(a) - \left(A(\eta) - A(\eta_m)\right)}.
\end{equation}
To verify Assumption~\ref{assum:holderian_error_bound}, we need to upper bound their variance, which we shall write as,
\[
    \Var_{a \sim \pi_{\eta_m}} \left[ \dfrac{\pi_\eta(a)}{\pi_{\eta_m}(a)} \right] = e^{2\left(A(\eta_m) - A(\eta)\right)} \Var_{a \sim \pi_{\eta_m}} \left[ e^{\left(\eta - \eta_m \right) t(a)} \right] \,.
\]
Now, computing the moment generating function (MGF) of the statistic $t(a) \in \R^q$
\begin{align*}
    M_t(s) = \E \left[ e^{s \cdot t(a)} \right] 
    = \int_{a} e^{s \cdot t(a)} e^{\eta_m \cdot t(a) -A(\eta_m)}h(a) \diff a 
    = e^{-A(\eta_m)} \int_{a} e^{(\eta_m + s) \cdot t(a)} e^{\eta_m \cdot t(a)}h(a) \diff a 
    = e^{A(\eta_m+s)-A(\eta_m)},
\end{align*}
the variance term may be written as
\[
    \Var_{a \sim \pi_{\eta_m}} \left[ e^{\left(\eta - \eta_m \right) t(a)} \right] = M_t(2(\eta-\eta_m)) - M_t^2(\eta-\eta_m) = e^{A(2\eta-\eta_m)-A(\eta_m)} - e^{2\left(A(\eta)-A(\eta_m)\right)}\,.
\]
This eventually leads us to
\begin{equation}
    \Var_{a \sim \pi_{\eta_m}} \left[ \dfrac{\pi_\eta(a)}{\pi_{\eta_m}(a)} \right] = e^{A(2\eta-\eta_m)+A(\eta_m)-2A(\eta)} - 1.
\end{equation}
We now discuss two cases that are used for discrete actions \citep{swaminathan2012} and continuous actions \citep{Kallus2018PolicyEA, zenati_counterfactual}.

\paragraph{Bounded sufficient statistic}
Supposing that there exists an upper bound $A$ such that $\Vert t(a) \Vert \leq A$, Cauchy-Schwartz inequality states that $\vert (\eta - \eta_m) \cdot t(a) \vert \leq \Vert \eta - \eta_m \Vert A$, which entails
\begin{align*}
    \Var_{a \sim \pi_{\eta_m}} \left[ \dfrac{\pi_\eta(a)}{\pi_{\eta_m}(a)} \right]  &  = e^{A(2\eta-\eta_m)+A(\eta_m)-2A(\eta)} - 1\\
    & = \dfrac{\int_a e^{(2\eta - \eta_m) \cdot t(a)} h(a) \diff a \int_a e^{\eta_m \cdot t(a)} h(a) \diff a}{\left(\int_a e^{\eta \cdot t(a)} h(a) \diff a \right)^2} - 1  \\
    &= \dfrac{\int_a e^{(\eta - \eta_m) \cdot t(a)} e^{\eta  \cdot t(a)} h(a) \diff a \int_a e^{(\eta_m - \eta) \cdot t(a)} e^{\eta \cdot t(a)} h(a) \diff a}{\left(\int_a e^{\eta \cdot t(a)} h(a) \diff a \right)^2} -1 \\
    &\leq e^{ \Vert \eta - \eta_m \Vert A} -1\,.
\end{align*}
Assuming that the parameter space is compact, i.e, $\max_{\eta,\eta'} \|\eta-\eta'\| \leq D$, there exists a constant $C$ that depends on $A$ and $D$ such that, this may be further upper-bounded as
\[
     \Var_{a \sim \pi_{\eta_m}} \left[ \dfrac{\pi_\eta(a)}{\pi_{\eta_m}(a)} \right]  \leq C \Vert \eta - \eta_m \Vert \,.
\]
Therefore, Assumption~\ref{assum:holderian_error_bound} is implied by
\[
    \gamma C \Vert \eta - \eta_m \Vert^2  \leq (L(\theta) - L(\theta^*))^{2\beta} \,.
\]
The latter is implied by a local version of strong convexity for $\beta = 1/2$ \citep{daspremont21}, and holds with $\gamma = C^{-1} D^{-2}$ for $\beta = 0$. 

\begin{exmp} \label{ex:expfamily}
For discrete actions $\cA = \{a_1, \dots, a_K \}$, we consider, as in \citep{swaminathan2012} and \citep{faury2020distributionally}, policies where given a context x, probabilities $p_i(x)$ of sampling an action $a_i$ are given by
\begin{equation}
    p_i(x) = \frac{\exp(\theta^\top \phi(x,a_i))}{\sum_{j=1}^K \exp(\theta^\top \phi(x,a_j))}.
\end{equation}
The function $\phi$ is typically a feature map associated to a kernel in a RKHS. In this case, the natural parameter $\eta$ and the sufficient statistic $t(a)$ may be written as
\begin{equation}
    \eta = \begin{bmatrix}
\log(\frac{p_1}{p_K}) \\
\vdots \\
\log(\frac{p_{K-1}}{p_K}) \\
0
\end{bmatrix}
\quad 
t(a) = \begin{bmatrix}
\mathbb{1}\{ a = a_1 \} \\
\vdots \\
\mathbb{1}\{ a = a_K \} 
\end{bmatrix} \,.
\end{equation}

\end{exmp}

\paragraph{Lognormal and Normal distributions}
For normal $\cN(\mu, \sigma^2)$ and lognormal $\text{Lognormal}(\mu, \sigma^2)$ distributions with fixed variance $\sigma^2$ as considered by  \citep{Kallus2018PolicyEA,zenati_counterfactual}, the normalizing constant writes $A(\eta)=\frac{\eta^2}{2}$, and we then obtain that:
\begin{equation*}
    A(2\eta-\eta_m)+A(\eta_m)-2A(\eta) = (\eta - \eta_m)^2 \,,
\end{equation*}
which gives:
\begin{equation*}
    \Var_{a \sim \pi_{\eta_m}} \left[ \dfrac{\pi_\eta(a)}{\pi_{\eta_m}(a)} \right] = e^{ \Vert \eta - \eta_m \Vert^2} - 1\,.
\end{equation*}
In that case, it is again possible for a bounded parameter space to linearize $e^{ \Vert \eta - \eta_m \Vert^2} - 1 \lesssim \Vert \eta - \eta_m \Vert^2$, consider losses that verify: for all $\eta$, there exists an optimal $\eta^*$ such that
\begin{equation}
    \gamma\Vert \eta_m - \eta^* \Vert^2 \leq \big( L(\eta_m) - L(\eta^*) \big)^{\beta}.
\end{equation}
Again, this holds generally for $\beta = 0$ and for locally strongly convex losses for $\beta = 1$.

\section{Experiment details}
\label{appendix:experiment_details}

\subsection{Code}

All the code to reproduce figures and tables is available in the following repository: \url{https://github.com/criteo-research/sequential-conterfactual-risk-minimization}.

\subsection{Empirical settings details}

\label{appendix:experiment_settings_details}



\paragraph{\benchpricing} The pricing application in \citep{demirer2019semi} considers a "personalized pricing" setting where given contexts $x$, prices $p$ (which are the actions) need to be predicted to maximize the revenue:

\begin{equation*}
    r(x, p) = p(a(x) - b(x)p + \epsilon)
\end{equation*}

where $\epsilon \sim \cN(0,1)$ and  $d = a(x)+b(x)p + \epsilon$ is akin to an unknown context-specifidemand function. The data generating process uses contexts $x \in [1,2]^k$ for $k>1$ a positive integer. Only $l<k$ dimensions however affect the demand, that is if we write $\bar x = \frac{1}{l}(z_1, \dots, z_l)$. The price $p$ is generated from a Gaussian logging policy $p \sim \cN(\bar x, 1)$ centered in $\bar x$. We consider in our example the quadratic functionnal $a(x) = 2x^2$ and $b(x)=0.6x$ as in the original paper. 

\paragraph{\benchpotential}  The advertising simulation in \citep{zenati_counterfactual} consists in predicting the potential $p \in ]0, + \infty [$ of a user that may be compared to their a priori responsiveness to a treatment. The potential is caused by an unobserved random group variable $g$ in $G$ (groups of "high" or "low" potential users in their responsiveness) that influences context $x$ of users. The goal is then to find a policy $\pi(a|x)$ that maximizes reward by adapting to an unobserved potential. The potentials are normally distributed conditionally on the group index, $p |g \sim \mathcal{N}(\mu_g, \sigma_g^2)$ where $\sigma_g = 0.5$ and $\mu_g = 1$ or $3$ for two groups. The observed reward~$-y$ is then a function of the action $a$ and the context~$x$ through the associated potential $p_x$ of the user $x$. The reward function mimics reward over the offline continuous bidding dataset in \citep{zenati_counterfactual} with the form:

\begin{align*}
    r_l(p_x, a) &= \left\{
    \begin{array}{ll}
        \frac{a}{p_x} & \mbox{if } a < p_x \\
        \frac{1}{2}(p_x - a) +1  & \mbox{else}
    \end{array}
    \right. \\
    r(p_x, a) &= \max(r_l(p_x, a), -0.1)
\end{align*}

The logging policy is a lognormal distribution as it is common in advertising applications \citep{bottou2012}. In particular, as in \citep{zenati_counterfactual}, $\pi_{\theta_0} = \text{Lognormal}(\mu, \sigma^2)$ where the mean $\exp(\mu + {\sigma^2}/{2}) = 2$ and the variance $(\exp(\sigma^2) -1) \exp(2 \mu + \sigma^2) = 1$. 

\paragraph{\benchyeast, \benchscene, \benchtmc}
We follow \cite{swaminathan2012}. We now recall briefly the setup. The problem is a binary multilabel classification with $|\cA| = 2^K$ potential labels.
All models are parametrized by $\pi_\theta( a | x) \propto \exp(\theta^\top (x \bigotimes a))$. The baseline (resp. skyline) is a supervised, full information model with identical parameter space than CRM methods trained on 5\% (resp. 100\%) of the training data.
Our main modification it to consider the class of probabilistic policies that satisfy Assumption \ref{assum:holderian_error_bound} by predicting actions in an Epsilon Greedy fashion \cite{sutton1998}): $\pi^{\epsilon}_\theta(a,x) = (1-\epsilon) \pi_\theta(a,x) + \epsilon/|\mathcal{A}|$ where $\epsilon=.1$.
The loss is the Hamming loss (number of incorrectly assigned labels - both false positives and false negatives in the action vector):
\begin{equation}
    L(\theta) = \frac{1}{n K} \sum_{i=1}^{n} \sum_{j=1}^{K} \mathbb{1}_{[y_i^j = a_i^j]}
\end{equation}
where $y_i^j$ (resp. $a_i^j$) is the $j$-th component of the label vector (resp. action vector) of line $i$. A uniform policy will thus evaluate at a loss of $.5$.

\subsection{Implementation details}

\label{appendix:experiment_implementation_details}

\paragraph{Counterfactual methods}

In this paragraph we start by detailing the non adaptive counterfactual risk minimization that we compare to in this work. 

\begin{algorithm}[h]
\SetAlgoLined
\KwIn{Logged observations $(x_{0,i}, a_{0,i}, y_{0,i}, \pi_{0,i})_{i = 1, \ldots, n_0}$, parameter $\lambda>0$}

 \For{$m=1$ to $M$}{
 Build $\cL_m$ from observations  $s_m$ using Eq. \eqref{eq:conservative_learning} \\
 Learn $\theta$ using Eq. \eqref{eq:learning_objective_scrm} \\
 Re-deploy the logging model $\theta_0$ and collect observations $s_{m+1}~=~(x_{m+1,i}, a_{m+1,i}, l_{m+1,i}, \pi_{m+1,i})_{i = 1, \ldots, n_{m+1}}$ \;
 }
 \caption{Counterfactual Risk Minimization}
 \label{alg:crm}
\end{algorithm}

We also provide the grid of hyperparameters for the $\lambda$ evaluated in CRM and SCRM methods $\lambda \in [1e-5, 1e-4, 1e-3, 1e-2, 1e-1]$.

\paragraph{Batch Bandits}

Let $k: (\cX \times \cA) \times (\cX \times \cA) \to \R$ be a bounded positive definite Kernel associated to a RKHS $\cH$, $\phi: \cX \times \cA \to \cH$ is the feature map such that $k(s,s') = \langle \phi(s), \phi(s')\rangle$ for any $s,s'\in \cX \times \cA$. Context-actions pairs are written as $s_{m,i} \defeq (x_{m,i},a_{m,i}) \in \cX \times \cA$ and $\cS_{m} \defeq \{s_{1,0},\dots, s_{n_m,m}\}$ denoting the history of all context-actions pairs seen up until the end of batch $m$. $K_m$ is the kernel matrix of all context-actions seen until the end of the batch $m\geq 1$. Eventually, $K_{\mathcal{S}}(s')$ is the kernel column vector $[k(s_1, s'), \dots, k(s_l, s')]^\top$ of size $|\mathcal{S}|=l$. $Y_m = [-y_{0, 1}, \cdots -y_{0,n_0}, \cdots -y_{m, 1}, \cdots -y_{m,n_m}]$ denotes the vector of concatenated rewards observed up until the end of the batch $m$.

At a batch $m$, a context $x_{m,i}$ is sampled for $i \in \{1, n_m \}$, and then to sample an action $a$, the following decision rule is applied:
\begin{equation}
    a \in \argmax_{a \in \mathcal{A}} \hat{q}_{m, i, a}.
\end{equation}
In batch Kernel UCB, $\hat{q}_{m, i, a}$ is defined as 
\begin{equation}
    \hat{q}_{m, i, a} = \hat{m}_{m, i, a} + \beta_m \hat{\sigma}_{m, i, a},
\label{eq:kucb}
\end{equation}
where 
\begin{align*}
    \hat{\mu}_{m, i, a} &= K_{\cS_{t-1}}\big((x_{m,i}, a)\big)^\top K_{m-1}^{-1} Y_{m-1} \\
    \hat{\sigma}_{m, i, a}^2 &= \frac{1}{\lambda} k\big((x_{m,i}, a), (x_{m,i}, a) \big) - \frac{1}{\lambda} K_{\cS_{m-1}}\big((x_{m,i}, a)\big)^\top K_{m-1}^{-1} K_{\cS_{m-1}} \big((x_{m,i}, a) \big),
\end{align*}
and $\beta_m$ is a theoretical parameter that is set to $\beta_m = \frac{1}{\sqrt{m}}$ in practical heuristics \citep{lattimore2019}.
In SBPE \citep{han2020sequential}, $\hat{q}_{m, i, a}$ is defined directly as 
\begin{equation}
    \hat{q}_{m, i, a} = K_{\cS_{t-1}}\big((x_{m,i}, a)\big)^\top K_{m-1}^{-1} Y_{m-1}.
\label{eq:sbpe}
\end{equation}
\begin{algorithm}[h]
\SetAlgoLined
\KwIn{Logged observations $(x_{0,i}, a_{0,i}, y_{0,i}, \pi_{0,i})_{i = 1, \ldots, n_0}$, $\lambda$ regularization and exploration parameters, $k$ the kernel function}
 initialization\\
 $K_\lambda = [k(s_{0, i},s_{0, j})]_{1 \leq i,j \leq n_0} + \lambda I, Y_0 = [-y_{0,i}]_{1 \leq i \leq n_0}$ \\
\For{$m=1$ to $M$}{
 \For{$i=1$ to $n_m$}{
 Observe context $x_{i, m}$ \\ 
 Choose $a_{i, m} \leftarrow \argmax_{a \in \mathcal{A}} \hat{q}_{m, i, a}$ using Eq. \eqref{eq:sbpe} or \eqref{eq:kucb} \\
}
 Observe losses $y_{i, m}$ for all $i$ in past batch $\{1, \dots, n_m \}$ \\
 Update $Y_m \leftarrow [-y_{0, 1}, \cdots -y_{0,n_0}, \cdots -y_{m, 1}, \cdots -y_{m,n_m}]$ \\
 Update the translated gram matrix $K_\lambda \leftarrow [k(s_{i,p},s_{j,p})]_{1 \leq i,j \leq n_p, 1\leq p \leq m} + \lambda I$ \\
 }
 
 \caption{Batch bandit - SBPE \citep{han2020sequential} and Kernel UCB \citep{valko_2013}}
 \label{alg:batch_bandit}
\end{algorithm}

SBPE \citep{han2020sequential} uses a linear modelling, therefore we used a linear kernel. For the Kernel UCB \citep{valko_2013} method, we used Gaussian and Polynomial kernels in our experiments. Note also that no regularization parameter $\lambda$ is used in SBPE so we set $\lambda = 0$ in our experiments, and for K-UCB we chose $\lambda$ in the grid $[1e0, 1e1, 1e2]$. 

Note in particular that we adapted the batch bandit baselines to the CRM setting by benefiting the initialization with the logged dataset to set the gram matrix $K_\lambda$ as well as the reward vector $Y_0$ with information from the logging data. This modification changes the original methods which take random actions at initializations.

Eventually, the baselines were carefully optimized using the Jax library (\url{https://github.com/google/jax}) to allow for just in time compilations of algebraic blocks in both methods and to maximize their scaling capacity.

\paragraph{RL baselines}

In order to compare our method to the two known off-policy online RL algorithm \verb|PPO| \citep{PPOref} and \verb|TRPO| \citep{TRPOref}, we do the following:
\begin{enumerate}
    \item we use the \verb|stable_baselines3|\citep{stable-baselines3} library for the implementation. When necessary we call multiple times the model \verb|PPO| or \verb|TRPO|, to have buffer size of geometrical increase.
    \item we initialize the \verb|ActorCriticPolicy| with a simpler MLP model having only one layer with output dimension of 1, (with argument \verb|net_arch= [1]|, that is mathematically the same modelling as in CRM and SCRM baselines).
    \item At the initial step only and to enable a fair comparison with counterfactual methods using a logging dataset, we pretrain the RL policies to imitate the actions sampled from the logging policy: we process by multiple step of the Adam optimizer, minimizing a loss being the sum of 2 terms:
    \begin{itemize}
        \item a MSE term between the sampled action of the \verb|ActorCriticPolicy| for the contexts in the $n_0$ instances, and the actions sampled by the logging policy.
        \item the ENTROPY term guaranteeing to keep a minimum of exploration in order to initialize the RL algorithm ($-\sum p_i \log(p_i)$)
    \end{itemize}
    \item we combine the 2 last terms with a linear combinaison with hyperparameters being tuned a posteriori, i.e. $\text{LOSS}=\text{MSE}+\lambda \text{ ENTROPY}$ with the hyperparam $\lambda \in \{ .5,1,2,5, 10 \}$
\end{enumerate}

\section{Additional empirical results}

\label{appendix:additional_empirical_results}

\subsection{SCRM compared to CRM}

\label{appendix:scrm_vs_crm}

We provide here the additional plot in the \benchpricing \ setting.

\begin{figure*}[h]
\label{fig:continuous-scrm_vs_crm}
    \centering
\begin{subfigure}
  \centering
    \includegraphics[width=0.35\linewidth]{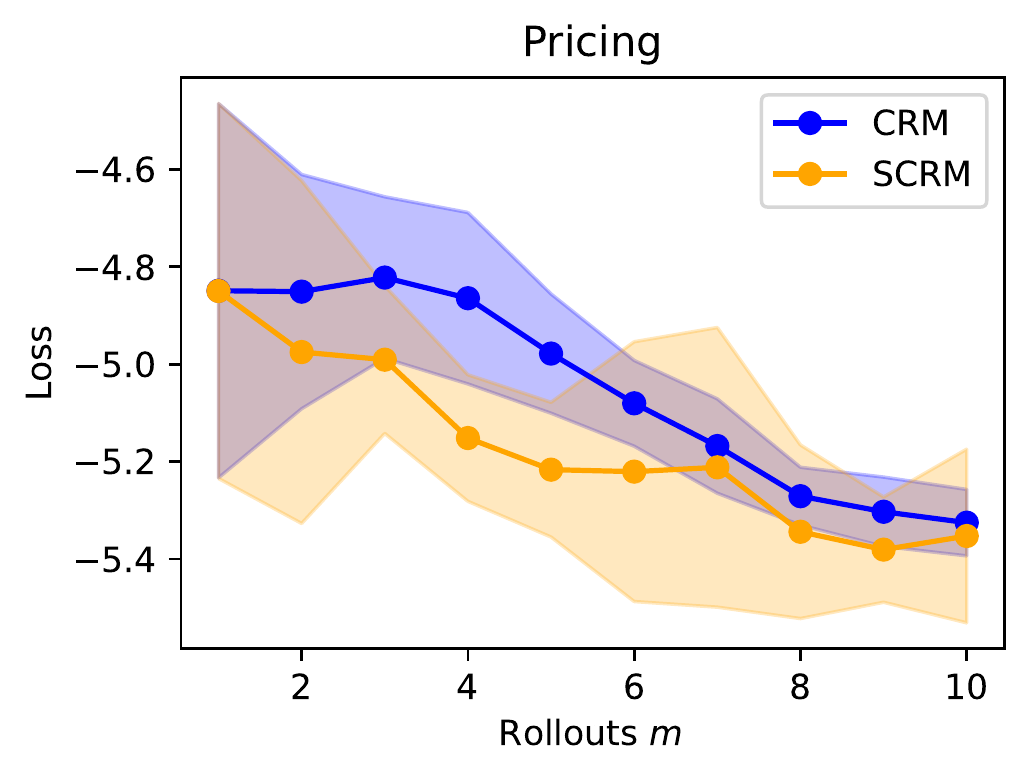}
\end{subfigure}%
\begin{subfigure}
  \centering
    \includegraphics[width=0.35\linewidth]{images/scrm_vs_crm/losses_scrm_vs_crm_advertising.pdf}
\end{subfigure}

\caption{Test loss as a function of sample size on \benchpricing, \benchpotential\ (from left to right).}
\end{figure*}

\subsection{Evaluation of IPS-IX}

\label{appendix:evaluation_ips_ix}
We provide here the plots for the whole setting considered in policy evaluation with IPS-IX.

\begin{figure*}[h]
    \centering
    \includegraphics[width=\textwidth]{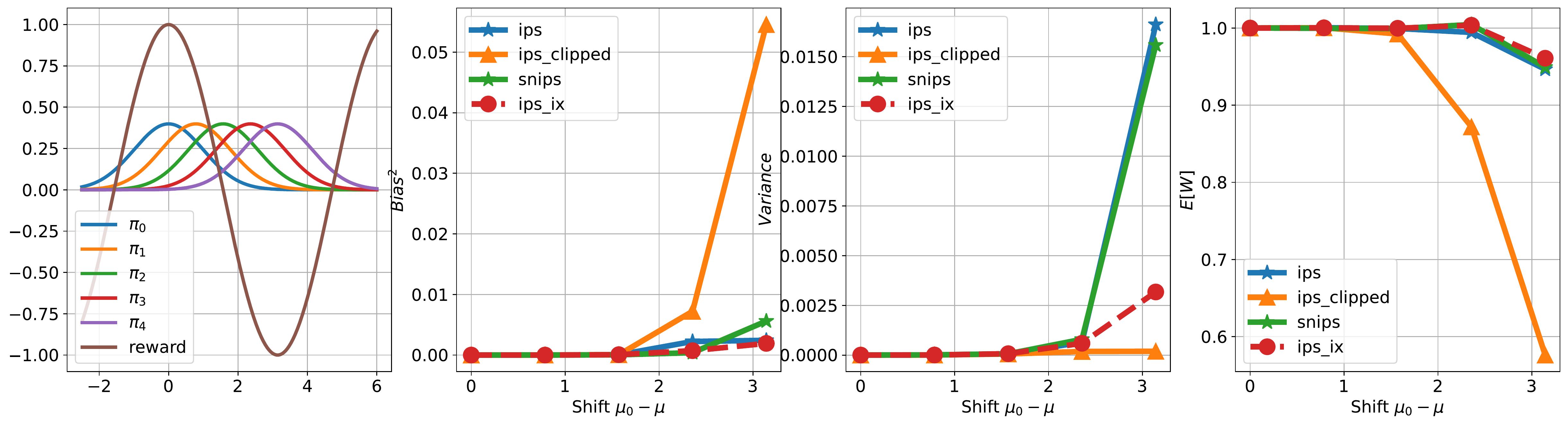}       
    \caption{Comparison of IPS estimators on a Cosine reward and series of shifted Gaussian policies. Setup (left), Bias (middle left), Variance (middle right), Average IPS weight (right). IPS-IX shows a low bias and compares favorably to IPS and SNIPS in terms of variance.}
    \label{fig:compare_estimators_complete}
\end{figure*}

\subsection{Exploration/Exploitation tradeoff}

\label{appendix:exploration_exploitation_tradeoff}

In this part we give the details used for the experiment described in Section \ref{sec:details_scrm}. We consider again Example \ref{example:gaussian_policies} with the Gaussian parametrized policies $\pi_\theta = \mathcal{N}(\theta, \sigma^2)$ and a loss $l_t(a)= (a-y_t)^2-1$ where $y_t \sim \mathcal{N}(\theta^*, {\sigma^*}^2)$ with $\sigma^*=0.3$. Recall that $\pi_{\theta_0} = \cN(\theta_0, \sigma)$.
We consider a grid of $\sigma \in [0.1, 0.3, 1, 3]$ and consider $\theta^*=1$. Our experiment aims at illustrating the influence of sequential exploration that is an important detail of the SCRM and CRM principles.